\documentclass{article}

\usepackage{microtype}
\usepackage{graphicx}
\usepackage{subfigure}
\usepackage{booktabs} 

\usepackage{hyperref}

\usepackage[accepted]{icml2024}

\usepackage{amsmath}
\usepackage{amssymb}
\usepackage{mathtools}
\usepackage{amsthm}

\usepackage[capitalize,noabbrev]{cleveref}

\theoremstyle{plain}
\newtheorem{theorem}{Theorem}[section]

\newtheorem{lemma}[theorem]{Lemma}

\theoremstyle{definition}

\theoremstyle{remark}

\usepackage[textsize=tiny]{todonotes}

\icmltitlerunning{Bayesian Power Steering}

\begin{document}

\twocolumn[
\icmltitle{Bayesian Power Steering: An Effective Approach for \\
Domain Adaptation of Diffusion Models}



\icmlsetsymbol{equal}{*}

\begin{icmlauthorlist}
\icmlauthor{Ding Huang}{yyy}
\icmlauthor{Ting Li}{yyy} 
\icmlauthor{Jian Huang}{yyy}
\end{icmlauthorlist}

\icmlaffiliation{yyy}{Department of Applied Mathematics, The Hong Kong Polytechnic University, Hong Kong, China}


\icmlcorrespondingauthor{Ting Li}{tingeric.li@polyu.edu.hk}
\icmlcorrespondingauthor{Jian Huang}{j.huang@polyu.edu.hk}

\icmlkeywords{Machine Learning, ICML}

\vskip 0.3in
]



\printAffiliationsAndNotice{}

\begin{abstract}
    We propose a Bayesian framework for fine-tuning large diffusion models with a novel network structure called Bayesian Power Steering (BPS) \footnote[2]{ Code and models are available at \url{https://github.com/DingDing33/BPS-v1-1}.}. We clarify the meaning behind adaptation from a \textit{large probability space} to a \textit{small probability space} and explore the task of fine-tuning pre-trained models using learnable modules from a Bayesian perspective. BPS extracts task-specific knowledge from a pre-trained model's learned prior distribution. It efficiently leverages large diffusion models,  differentially intervening different hidden features with a head-heavy and foot-light configuration. Experiments highlight the superiority of BPS over contemporary methods across a range of tasks even with limited amount of data. Notably, BPS attains an FID score of 10.49 under the sketch condition on the COCO17 dataset.

\end{abstract}

\section{Introduction}\label{introduction}

The advent of diffusion models \citep{ho2020denoising, song2020score} and their extensions \citep{song2020denoising, nichol2021improved, huang2023conditional}, has enabled effective learning of intricate probability measures for diverse data types, including images \citep{ho2022cascaded, rombach2022high, saharia2022photorealistic, ho2022cascaded}, audio \citep{kong2020diffwave}, and 3D bioimaging data \citep{luo2021diffusion, poole2022dreamfusion, shi2023mvdream, pinaya2022brain}. For these generative models, the quantity of training data plays a crucial role in influencing both the precision of probability measure estimation and the generalization capacity, enabling them to extrapolate effectively within the probability space. Particularly, in computer vision, significant efforts have been devoted to developing large-scale diffusion models. For instance, the Stable Diffusion (SD, \citet{rombach2022high}) is trained utilizing the LAION-5B dataset \citep{schuhmann2022laion}, a large publicly available text-image dataset, with a staggering magnitude of 585 billion.

\begin{figure}[ht]
    \vskip 0.05in
    \begin{center}
    \centerline{\includegraphics[width=\columnwidth]{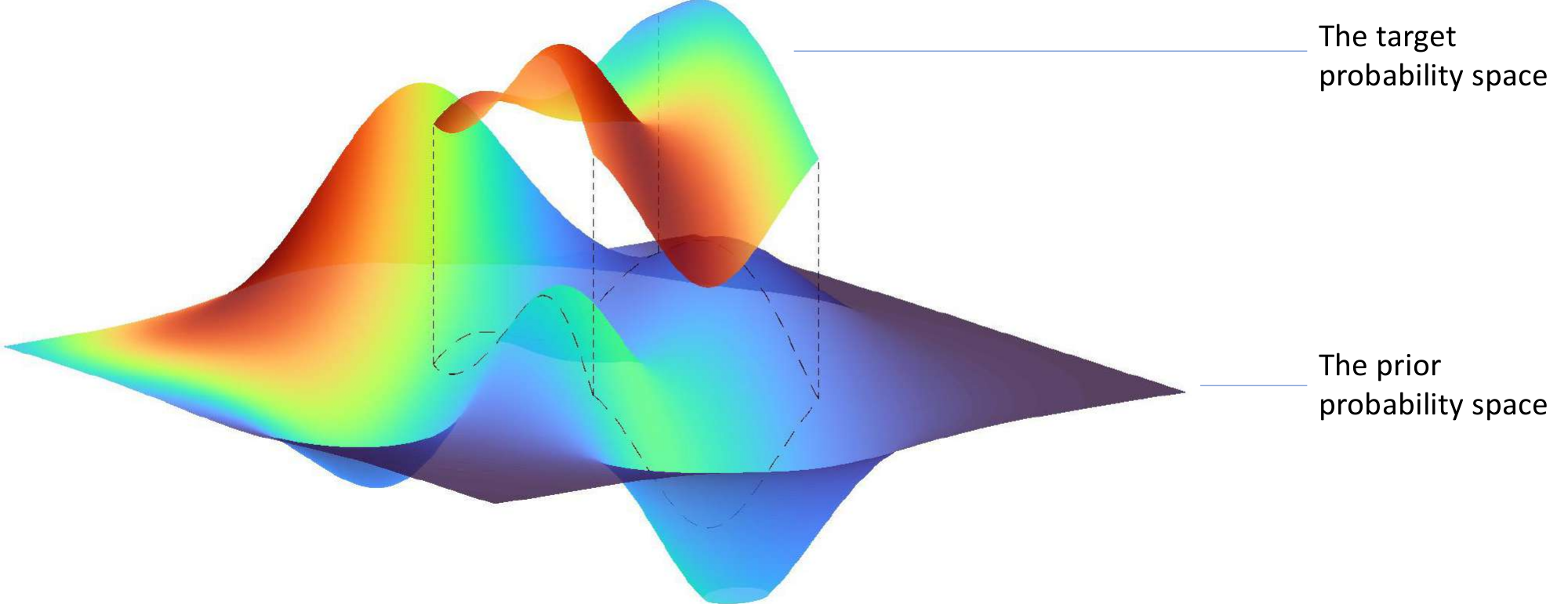}}
    \caption{The relationship between the probability spaces learned by the pre-trained model (the prior probability space or \textit{large probability space}) and the target task (the target probability space or \textit{small probability space}). Different colors represent different densities at various locations on the manifold.}
    \label{latent_fig}
    \end{center}
    \vskip -0.25in
\end{figure}

The diversity and abundance of data contributes to the exceptional generative capabilities of large-scale models. They can capture intricate details in the large probability space of images. However, numerous data-driven modeling tasks typically focus on a specific subset of the entire image space, as illustrated in Figure \ref{latent_fig}. Moreover, the size of the training data available for these tasks is considerably smaller compared to expansive datasets such as LAION-5B. This limitation is especially evident in user-customized scenarios, where the available number of samples provided by users is often limited. Consequently, learning this \textit{smaller probability space} poses a significant challenge.

Pre-trained large models, which encapsulate information across the \textit{ large probability space} and exhibit exceptional generalization abilities, offer a potential solution to the challenge posed by limited training data sizes. This prompts us to ask the question: can the judicious deployment of such pre-trained large-scale models, in conjunction with carefully curated specialized datasets, effectively facilitate the transition from a \textit{large probability space} to a \textit{small probability space}?

A related topic is \textbf{domain adaptation for generative tasks}. Originating from the field of transfer learning, this concept has gradually extended its application to generation tasks, particularly aimed at transferring a broader knowledge base acquired from publicly available data to the task-specific distribution \citep{li2021large, harder2022pre,lyu2023differentially, kurakin2023harnessing}. Especially in privacy protection and medical data domains, with a specific focus on creating infinite amounts of synthetic data to maximise the downstream prediction performance \citep{sagers2023augmenting,ghalebikesabi2023differentially, li2023meticulously, tang2024differentially}. Another notable application lies in customizing scenarios for users. Textual inversion \citep{gal2022image}, DreamBooth \citep{ruiz2023dreambooth}, LayoutDiffuse \citep{cheng2023layoutdiffuse} and StyleDrop \citep{sohn2023styledrop}  are exemplary techniques to capture the specific domain (semantic concepts, patterns, styles, etc.), leveraging a small set of user-provided example images.

Other efforts have been devoted to \textbf{conditional control for text-to-image diffusion models} for achieving customized image generation. Text-based control methods, as explored by  \citet{brooks2023instructpix2pix, hertz2022prompt, gafni2022make, kawar2023imagic, parmar2023zero, chefer2023attend, orgad2023editing}; and \citet{ couairon2022diffedit}, are centered around the adjustment of prompts, manipulation of CLIP features, and modification of cross-attention mechanisms. The methods of \cite{ho2022classifier, hong2023improving, nichol2021glide, brack2023sega} combine diverse prompts to achieve desired outcomes. For control conditions other than text, \citet{dhariwal2021diffusion} introduce a time-dependent classifier, which requires a separate training procedure and leads to a performance decline \citep{ghalebikesabi2023differentially} due to a decreasing signal-to-noise ratio over time. Concurrent endeavors including T2I-Adapter \citep{mou2023t2i} and ControlNet \citep{zhang2023adding} employ fine-tuning techniques on large-scale diffusion models, which enable precise spatial control over image generation and establish the state of the art (SOTA).

In this paper, we first mathematically formulate the problem,
proposing a framework for transition from a \textit{large probability space} to a \textit{small probability space} through its transformation into a conditional generative problem. To tackle this problem, we propose a Bayesian fine-tuning framework along with a novel network structure called Bayesian Power Steering (BPS), designed to shift the pretrained diffusion system towards the data support specified by the given conditions.

Our extensive experiments demonstrate that
BPS, with significantly reduced amounts of data, generates samples that align with the support of the dataset and achieves comparable generative quality to large-scale models. Our ablation studies confirm the robustness and scalability of BPS across datasets of varying sizes. Experiments validate the effectiveness of the proposed model,
the necessity of its components, and , and provide a comparative analysis against strong conditional image generation baselines. Notably,the proposed method surpasses state-of-art methods, achieving the FID score of $10.49$ with sketch condition on the COCO17 dataset \citep{lin2014microsoft}.

\section{Preliminary: Stable Diffusion}
Throughout this paper, we adopt the Stable Diffusion \citep{rombach2022high} as the pre-trained large-scale model.
Initially, the image data $X_0 \in \mathbb{R}^{d_{\text{img}}}({d_{\text{img}}} =  512^2 \times 3)$ is mapped to a lower-dimensional latent probability space $\mathcal{Z} := (\Omega, \mathfrak{F}, \mathcal{P}) \subset \mathbb{R}^d (d= 64^2 \times 4)$ via the pre-trained autoencoder  \citep{esser2021taming}, while related text prompts are encoded as $C_{\text{text}} \in \mathbb{R}^{k_{\text{text}}}$ by CLIP \citep{radford2021learning}. Subsequently, the diffusion process  \citep{ho2020denoising} is employed to generate representations of the image in the latent space with condition $C_{\text{text}}$. Finally, the image is reconstructed using the pre-trained decoder.

The forward diffusion process in the latent space $\mathbb{R}^d$ is expressed as $\{Z_t\}_{t=0}^T$  with $t \in [T] := \{1, ..., T\}$:
\begin{align} \label{s2_forward}
    Z_t = \sqrt{\Bar{\alpha}_t} Z_0 + \sqrt{1-\Bar{\alpha}_t} \boldsymbol{\eta}, \text{ }\boldsymbol{\eta} \sim \mathcal{N}(\mathbf{0}, \mathbf{I}_{d}),
\end{align}
where $Z_0  \in \mathcal{Z}$ denotes the generative object, and $\{\Bar{\alpha}_t\}_{t=1}^T$ is a strictly decreasing sequence within the interval $(0,1)$. In the reverse process, conditional modeling is accomplished by manipulating the direction of the score function with conditions. The backward process $\{\Tilde{Z}_t^{C_{\text{text}}}\}_{t=1}^T$ starts from $\Tilde{Z}_T \sim \mathcal{N}(\mathbf{0},\mathbf{I}_{d})$ with the following iteration:
\begin{align} \label{s2_ddpm}
    \Tilde{Z}_{t-1}^{C_{\text{text}}} =& {\frac{1}{1-\beta_t}}\left(\Tilde{Z}_t^{C_{\text{text}}}- \frac{\beta_t}{\sqrt{1-\Bar{\alpha}_t}} \boldsymbol{\epsilon}^*(\Tilde{Z}_t^{C_{\text{text}}},t, C_{\text{text}})\right) \nonumber \\
    &+ \sigma_t \boldsymbol{\eta},
\end{align}
, here $\beta_t := 1-{\bar{\alpha}_{t}}/{\bar{\alpha}_{t-1}}$, $\sigma_t :=\frac{1-\bar{\alpha}_{t-1}}{1-\bar{\alpha}_t} \beta_t$. According to Tweedie's formula \citep{efron2011tweedie},
the denoise function $\boldsymbol{\epsilon}^*(\mathbf{z}_t,t,\mathbf{c}_{\text{text}})$ can be expressed as
\begin{align} \label{s2_denoise}
    \boldsymbol{\epsilon}^*(\mathbf{z}_t, t, {\mathbf{c}_{\text{text}}}) :&= - \sqrt{1-\bar{\alpha}_{t}} \nabla \log p(Z_t= \mathbf{z}_t \mid C_{\text{text}} = {\mathbf{c}_{\text{text}}}) \nonumber \\
    &= \mathbb{E}[\boldsymbol{\eta} \mid Z_t = \mathbf{z}_t, C_{\text{text}} = {\mathbf{c}_{\text{text}}}].
\end{align}

The denoise function  is parameterized through a neural network, denoted as $\boldsymbol{\epsilon_{\hat{\theta}}}$. As demonstrated in equation (\ref{s2_ddpm}), precise control over the sampling process can be attained by conducting fine-tuning on the pre-trained denoise function $\boldsymbol{\epsilon_{\hat{\theta}}}$.

\section{General Setup}

Our objective is to fine-tune the pre-trained denoise function $\boldsymbol{\epsilon_{\hat{\theta}}}$ with learnable modules, aiming to extract a \textit{small probability space} from a \textit{large probability space}.

\subsection{Problem Formulation} \label{sec31}
We first establish the precise mathematical definitions for both  \textit{large} and \textit{small latent probability spaces}. Let $Z_0$ defined in the latent probability space $\mathcal{Z}:=(\Omega, \mathfrak{F}, \mathcal{P})$ be the generative target of the pre-trained model. Our focus lies on a \textit{small latent probability space} residing within a non-zero measurable set $\Delta \in \mathfrak{F}$ (Fig.\ref{latent_fig}). This space corresponds to the trace of $\mathcal{Z}$ on $\Delta$ and is denoted as $\mathcal{Z}_\Delta := (\Delta, \Delta \cap \mathfrak{F}, \mathcal{P}_\Delta)$.  Specifically, the target probability measure is defined by $ \mathcal{P}_\Delta(E) := \mathcal{P}(E)/\mathcal{P}(\Delta)$, for all $E \in \Delta \cap \mathfrak{F}$.

According to the following lemma, the target domain $\Delta$ can be described using a suitable condition $\mathbf{c} \in \mathbb{R}^k$.

\begin{lemma}\citep{chung2001course} \label{lemma}
    If $\Delta \in \mathfrak{F}$, then there exists some measurable function $\boldsymbol{\psi}(\cdot)$ such that $\Delta(\omega) = \boldsymbol{\psi}(Z_0)$ for any $\Delta(\omega) := \omega  \in \Delta \cap \mathfrak{F}$.
\end{lemma}

By introducing a function $\boldsymbol{\psi}: \Omega \rightarrow \mathbb{R}^k$ to describe the relationship $\mathbf{c} = \boldsymbol{\psi}(\mathbf{z}_0)$, $\Delta$ can be defined as the union of sets $\Delta = \cup_{\mathbf{c} \in \boldsymbol{\psi}(\Delta)}\{\mathbf{z}_0: \boldsymbol{\psi}(\mathbf{z}_0) = \mathbf{c}\}$. Consequently,  when the data $\mathbf{z}_0 \in \Delta$ aligns with suitable conditions ${\mathbf{c}}$, the task of learning a ``small distribution" is formulated as learning a probability measure of $Z_0 \mid C$, where $C$ is a random variable defined in $\mathcal{C} := (\boldsymbol{\psi}(\Delta), \boldsymbol{\psi}(\mathfrak{F}), \mathcal{P}^\prime) \subseteq \mathbb{R}^k$.

The form of conditions can vary widely. For example, in Text-to-Image Diffusion Models, textual descriptions, denoted as $C_{\text{text}}$, are employed to guide the generative process. However, given that a single image can correspond to multiple descriptions, establishing a direct mapping $\psi$ from image to text alone is not practical. To address this, additional descriptors $C_{\text{add}}$, such as edge, depth, or attention information, can be incorporated based on the specific goals of the model. These supplementary controls, which can be numerous, aid in pinpointing the desired target support $\Delta$. Consequently, the condition in this context is represented as a $k$-dimensional vector $C :=(C_{\text{text}}^\top, C_\text{add}^\top)^\top$, integrating both textual and additional descriptive elements.

The proposed BPS method, which is formally introduced in Section \ref{BPS}, is based on the above formulation for fine-tuning the pre-trained denoise function $\boldsymbol{\epsilon_{\hat{\theta}}}$  using independent paired samples from $\mathcal{X} \times \mathcal{C}$.

\begin{figure*}[ht]
    \vskip 0.07in
    \begin{center}
\centerline{\includegraphics[width=0.95\textwidth, height=1.6 in]{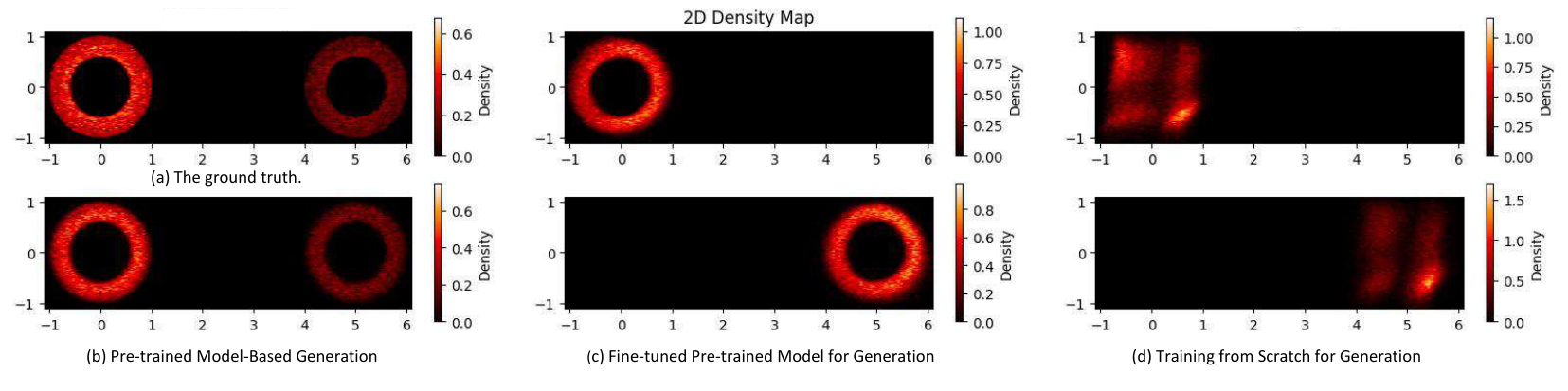}}
    \vskip -0.1in
    \caption{The simulation of 2D measure. The pre-trained model is trained with 500K samples, and the generative results are shown in Fig. (b). In Figures (c) and (d), labeled subsets of 100 samples were utilized in the experiments. Fig. (c) illustrates the generation outcomes achieved through fine-tuning the pre-trained model with 2.5K epochs, whereas Fig. (d) displays the results obtained by training the same model from scratch with 20K epochs. The above legends represent the sampling of 5K samples on either the simulated or real measure.}
    \label{1d-dist}
    \end{center}
    \vskip -0.25in
\end{figure*}

\subsection{Bayesian Formulation}

Suppose condition $(\mathbf{c}_{\text{text}}, \mathbf{c}_{\text{add}})$ provides a detailed characterization of the target domain, then our primary objective is to learn the integrated denoise function, denoted as
\begin{align*}  
    \bar{\boldsymbol{\epsilon}}^* (\mathbf{z}_t,t, \mathbf{c}_{\text{text}}, \mathbf{c}_{\text{add}})
     :=  \mathbb{E}[\boldsymbol{\eta} \mid  \mathbf{z}_t, \mathbf{c}_{\text{text}}, \mathbf{c}_{\text{add}}] .
\end{align*}
To take advantage of the pretrained model, we bridge $ \bar{\boldsymbol{\epsilon}}^* (\mathbf{z}_t,t, \mathbf{c}_{\text{text}},\mathbf{c}_{\text{add}}) $ and $\boldsymbol{\epsilon}^*(\mathbf{z}_t,t, \mathbf{c}_{\text{text}}),$ using Tweedie's formula \citep{efron2011tweedie} and Bayes' theorem as follows. First, Tweedie's formula gives
\begin{align*} 
\bar{\boldsymbol{\epsilon}}^* (\mathbf{z}_t,t, \mathbf{c}_{\text{text}}, \mathbf{c}_{\text{add}})     = -{\sqrt{1-\bar{\alpha}_t}}\, \nabla \log p(\mathbf{z}_t \mid \mathbf{c}_{\text{text}}, \mathbf{c}_{\text{add}}). 
\end{align*}

By Bayes' theorem, we have
\begin{align}
 \label{bayes1}
p( \mathbf{z}_t \mid \mathbf{c}_{\text{text}}, \mathbf{c}_{\text{add}})
= \frac{ p(\mathbf{c}_{\text{add}} \mid  \mathbf{z}_t, \mathbf{c}_{\text{text}}) }
{p(\mathbf{c}_{\text{add}}\mid \mathbf{c}_{\text{text}}) }   p(\mathbf{z}_t\mid \mathbf{c}_{\text{text}}).
\end{align}
Taking the logarithm across (\ref{bayes1}), differentiating with respect to $\mathbf{z}_t$,
and then using the definitions of $\bar{\boldsymbol{\epsilon}}^*(\mathbf{z}_t,t, \mathbf{c}_{\text{text}}, \mathbf{c}_{\text{add}})$ given above
and $\boldsymbol{\epsilon}^*(\mathbf{z}_t,t, \mathbf{c}_{\text{text}})$ from Equation (\ref{s2_denoise}),
we have

\vspace{-0.2in}
\begin{align} \label{s3_idea}
   & \bar{\boldsymbol{\epsilon}}^*(\mathbf{z}_t,t, \mathbf{c}_{\text{text}}, \mathbf{c}_{\text{add}})\\
   & =-{\sqrt{1-\bar{\alpha}_t}} [\nabla \log p(\mathbf{c}_{\text{add}} \mid  \nonumber
 \mathbf{z}_t,  \mathbf{c}_{\text{text}})
    + \nabla \log p( \mathbf{z}_t \mid  \mathbf{c}_{\text{text}})] \nonumber \\
    &= -{\sqrt{1-\bar{\alpha}_t}} \nabla \log p( \mathbf{c}_{\text{add}} \mid  \mathbf{z}_t,  \mathbf{c}_{\text{text}}) + \boldsymbol{\epsilon}^*(\mathbf{z}_t,t, \mathbf{c}_{\text{text}}). \nonumber
 \end{align}
 Detailed derivations are deferred to Appendix \ref{app_derivation}. Below, we denote
\begin{align}
\label{ps1}
M(\mathbf{z}_t, t, \mathbf{c}_{\text{text}}, \mathbf{c}_{\text{add}})= -{\sqrt{1-\bar{\alpha}_t}} \nabla \log p(\mathbf{c}_{\text{add}} \mid \mathbf{z}_t,  \mathbf{c}_{\text{text}}).
\end{align}

The integrated function $\bar{\boldsymbol{\epsilon}}^* $ can be interpreted as the \textit{posterior denoise function} corresponding to the \textit{prior denoise function} ${\boldsymbol{\epsilon}}^*$ for the pretrained model. It is obtained by combining the pretrained denoise function ${\boldsymbol{\epsilon}}^*$ with a time-dependent ``steering gear" $M$.

We emphasizes the role of the learnable modules as time-dependent ``steering gears"
$\{M(\mathbf{z}_t, t, \mathbf{c}_{\text{text}}, \mathbf{c}_{\text{add}}), t=1,\ldots, T\}$ within the architecture, guiding the system towards directions with a high probability density of satisfying condition $C$.  However, such externally appended fine-tuning structure introduces the problem of high-dimensional input-output, thereby increasing the computational burden during training. To address it, we integrate the information of $\mathbf{c}_{\text{add}}$ into the pretrained denoising function by realizing this corollary in the feature space, discussed in the following subsection.

\subsection{Integration Strategy}
Our study utilizes a pretrained model $\boldsymbol{\epsilon_{\hat{\theta}}}$  based on the U-net architecture \citep{ronneberger2015u}. This architecture consists of an encoder (E.), a middle block (MB.), a skip-connected decoder (D.), and skip-connections between the encoder and decoder (E-D.). These components sequentially extract information from the input, yielding distinctive levels of feature space.

\begin{table}[ht] 
    \caption{Sites of modular integration with pre-trained models.}
    \label{integration-table}
    \vskip 0.15in
    \begin{center}
    \begin{small}
    \begin{sc}
    \begin{tabular}{lcccr}
    \toprule
    Mode & E. & MB. & D. & E-D. \\
    \midrule
    ALL    & $\surd$ & $\surd$ & $\surd$ & $\surd$ \\
    EMD    & $\surd$ & $\surd$ & $\surd$ & $\times$ \\
    E    & $\surd$ & $\times$ & $\times$ & $\times$ \\
    EM    & $\surd$ & $\surd$ & $\times$ & $\times$ \\
    D    & $\times$ & $\times$ & $\surd$ & $\times$ \\
    MD    & $\times$ & $\surd$ & $\surd$ & $\times$ \\
    E-D    & $\times$ & $\times$ & $\times$ & $\surd$ \\
    ME-D    & $\times$ & $\surd$ & $\times$ & $\surd$ \\
    M    & $\times$ & $\surd$ & $\times$ & $\times$ \\
    \bottomrule
    \end{tabular}
    \end{sc}
    \end{small}
    \end{center}
    \vskip -0.17in
\end{table}

Based on the insights from equation \ref{s3_idea}, we explore potential schemes for integrating residual structures across various hierarchical levels of the feature space, as outlined in Table \ref{integration-table}. Notably, mode ME-D is adopted in ControlNet \citep{zhang2023adding} and mode EM is adopted in T2I-Adapter \citet{mou2023t2i}. We conduct a comparative analysis of the performance exhibited by different integration modes, considering the variables $t$ and $\mathbf{c}_{\text{add}}$ as inputs, in following generation tasks.

\textbf{Generation of multimodal 2D data.} We use this example to demonstrate the formulation in subsection \ref{sec31} and provide a simple illustration of the impact of conditional information on the denoising direction offset. As show in Fig. \ref{1d-dist}(a), the complete support comprises two rings, and the target domain $\Delta$ is a specified ring. We construct a U-net tailored for this vector data and pre-train a large-scale model. The generation results are presented in Fig. \ref{1d-dist}(b).

\begin{figure}[ht]
    \begin{center}
    \centerline{\includegraphics[height= 2.2 in, width=3.0 in ]{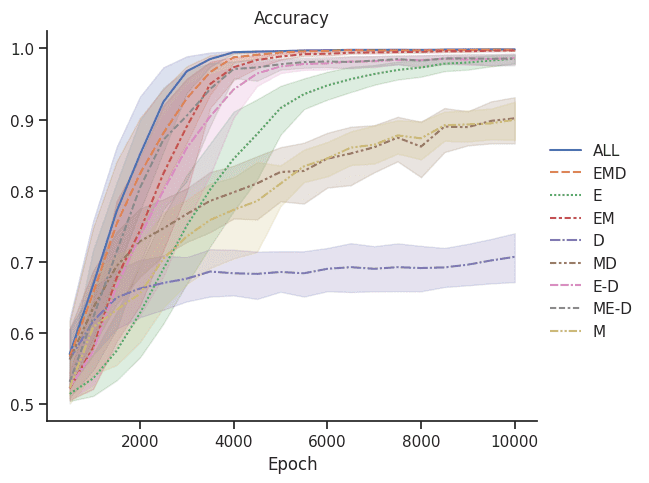}}
    \vspace{-0.1in}
    \caption{Classification accuracy of generated samples from distinct integration modes with $n=12$ training samples. For each mode, five experiments are conducted.
    }
    \label{1d-acc-fig}
    \vspace{-0.2in}
    \end{center}
\end{figure}

We quantify the effectiveness of integration modes by the accuracy of the support set to which the generated samples belong. According to Fig.\ref{1d-acc-fig}, the generative performance can be classified into three tiers. The first tier (ALL, EMD, and EM) achieves the highest accuracy and the fastest convergence rate. The second tier (E, EM, and ME-D) and the third tier (M, MD, and D) exhibit relatively worse performance. Furthermore, the comparison of mode E and EM (D and MD/E-D and ME-D) highlights the critical impact of injecting conditional information in the intermediate blocks. Moreover, the analysis of modes E, M, and D elucidates that introducing conditional information at different levels of the feature space leads to varying degrees of influence on the output gradient offset. Notably, injecting information at earlier stages exerts a more significant impact on the directional offset.

\begin{figure}[ht]
    \begin{center}
    \centerline{\includegraphics[width=\columnwidth]{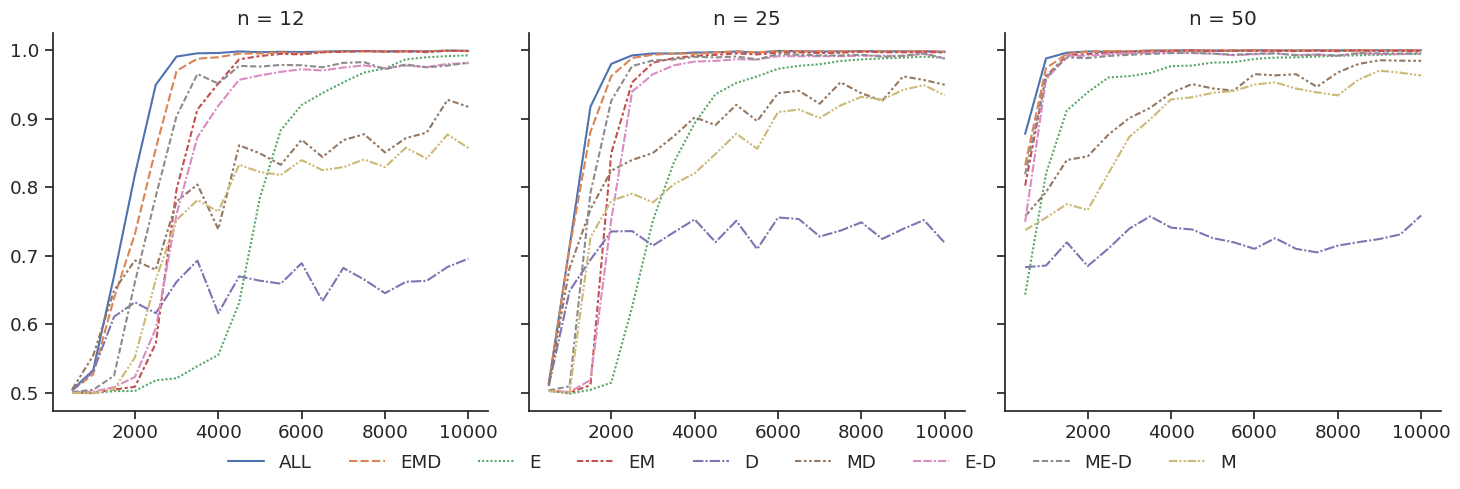}}
    \vskip -0.05in
    \caption{Robustness analysis of the integrated Models of distinct modes with respect to sample size and number of iterations. The horizontal coordinate is the number of epochs and the vertical coordinate is the accuracy.}
    \label{1d-n-fig}
    \end{center}
    \vskip -0.2in
\end{figure}

The robustness of various integration models concerning sample size and the number of iterations is shown in Fig.\ref{1d-n-fig}. The first tier modes (ALL, EMD, and EM) consistently demonstrate superior performance and convergence speed.

\textbf{Segmentation-to-image.} For this task, the performance is assessed using the ADE20K dataset \citep{zhou2017scene}. The conditioning fidelity is evaluated through Mean Intersection-over-Union (mIoU), and the state-of-the-art segmentation method OneFormer \citep{jain2023oneformer} achieves the mIoU of 0.58.

We utilize the pre-trained SD as the backbone and exam mode ALL, EMD, EM, MC-E, and MD. We generate images using segmentations from the ADE20K validation set and then feed the generative results to OneFormer for segmentation detection and computation of reconstructed IoUs. From Fig.\ref{1d-n-fig}, these modes exhibit superior performance in different tiers.

Table \ref{table-miou} presents the evaluation of semantic segmentation label reconstruction with mIoU scores. Mode ALL, EMD, and MD exhibit comparable performance. Notably, mode MD outperforms mode ME-D, highlighting the significant impact of the decoder component in complex tasks.

\begin{table}[h] 
    \vskip -0.1in
    \caption{Evaluation of semantic segmentation label reconstruction with Mean Intersection over Union (mIoU $\uparrow$).}
    \label{table-miou}
    \vskip 0.15in
    \begin{center}
    \begin{small}
    \begin{tabular}{lccccr}
    \toprule
    ALL & EMD & EM & ME-D & MD & BPS\\
    \midrule
    $0.351$ & $0.351$ & $0.350$ & $0.163$ & $0.240$  & $\mathbf{0.366}$ \\
    \bottomrule
    \end{tabular}
    \end{small}
    \end{center}
    \vskip -0.1in
\end{table}

\section{Bayesian Power Steering (BPS)} \label{BPS}
BPS incorporates a pre-trained SD model through the EMD integration mode to deal with task-specific conditions. As an instantiation of the Bayesian formulation in hierarchical levels of the feature space, BPS excels in both computational efficiency and domain recognition across multiple tasks.

\subsection{Overview}\label{sec-overall}
\begin{figure}[ht]
    \vskip 0.07in
    \begin{center}
    \centerline{\includegraphics[width=\columnwidth, height=4.7cm]{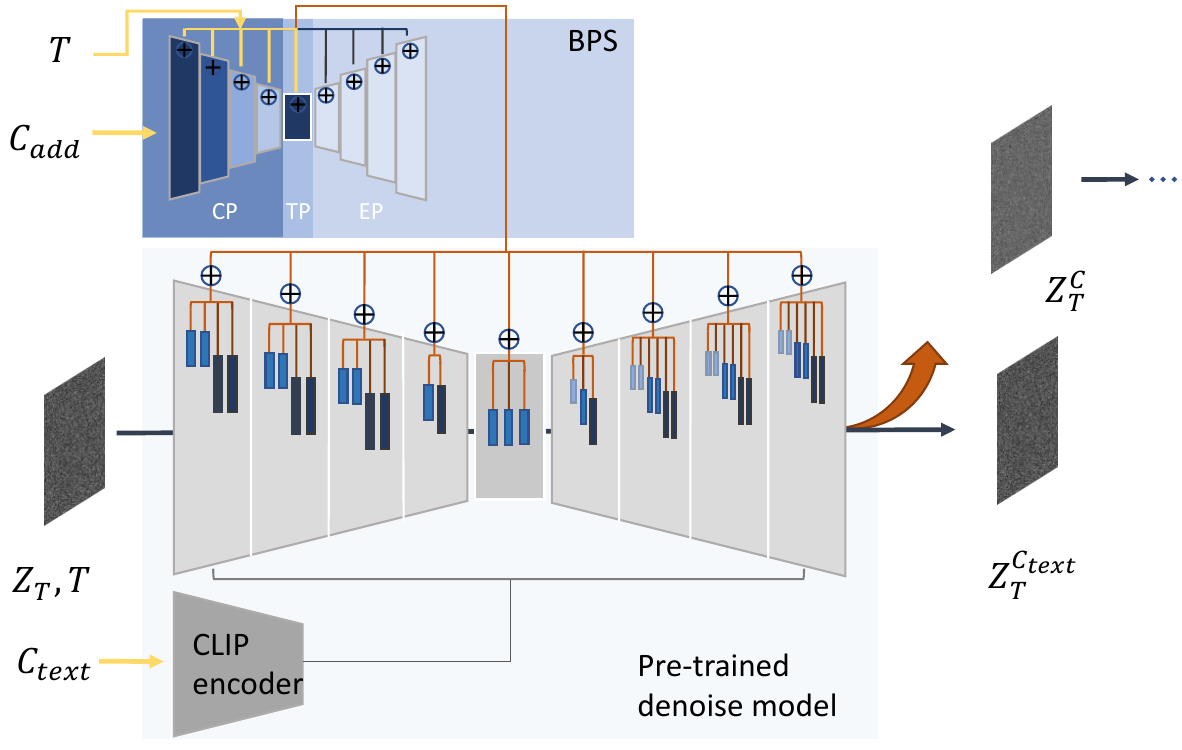}}
    \caption{Overview of the integration model. The gray dumbbell shapes represent the pre-trained model. The blue rectangles represent blocks with residual structures, and rectangles of the same color in the same scale represent interventions in the same functional unit sharing features from the BPS.}
    \label{overall}
    \end{center}
    \vskip -0.3in
\end{figure}

BPS takes time step information and additional conditions as inputs to perturb the latent features within the pre-trained model, thereby altering the denoising trajectory. Fig.\ref{overall} shows the main components of BPS: a contracting path (CP), a transition path (TP), and an expansive path (EP). These components are specifically designed to extract features at different scales, enabling adaptation to the diverse hidden features of the pre-trained model.

The dimension of the additional condition $\mathbf{c}_{\text{add}}$ is user-defined, and we take $512 \times 512$ for single image-type condition. Following the initial unshuffle module \citep{shi2016real}, we downsample $\mathbf{c}_{\text{add}}$ to $64 \times 64$. Subsequently, within the network, the scales for the extracted condition features range from $8 \times 8$ to $64 \times 64$. Through the components CP, TP, and EP, $8$, $1$, and $12$ condition features are extracted respectively, denoted as $\mathbf{v}_{\text{add}} = \{\mathbf{v}_{\text{add}}^{i} \mid i = 1,2,..., 21 \}$.

Note that the SD backbone contains 21 functional units, which refers to the smallest operational unit in the UNet structure and possess residual structures that serve as binding sites. The dimension of $\mathbf{v}_{\text{add}}$ aligns with the dimension of the latent feature $\mathbf{h}=\{ \mathbf{h}^{i,j} \mid i = 1,2,..., 21, i\text{-th unit, } j\text{-th block}\}$ within the functional unit of the UNet denoiser. The integration of BPS with the blocks across different units is illustrated in Fig.\ref{overall}. The process of feature extraction and fine-tuning latent features at different scales can be summarized as follows:
\begin{align}
    & \mathbf{v}_{\text{add}}= B_{\boldsymbol{\phi}}(t,  \mathbf{c}_{\text{add}}) \\
    & \hat{\mathbf{h}}^{i,j}=\mathbf{h}^{i,j}+\mathbf{v}_{\text{add}}^i, i= 1,2, ..., 21, \label{equ-combining}
\end{align}
where $B_{\boldsymbol{\phi}}$ represents BPS, which can be considered as the realization of steering gear $M$ defined in (\ref{ps1}) resttricted
to the feature space, $\boldsymbol{\phi}$ denotes the learnable parameters, and $\hat{\mathbf{h}}$ represents the disturbed latent feature. To train the integration model $\bar{\boldsymbol{\epsilon}}_{\hat{\boldsymbol{\theta}}, \boldsymbol{\phi}} (\mathbf{z},t, \mathbf{c}_{\text{text}}, \mathbf{c}_{\text{add}})$ that combines 
BPS and the SD backbone, we employ the following optimization process:
\begin{align}
    \mathcal{L}_{\boldsymbol{\phi}}=\mathbb{E}_{Z_0, t, \boldsymbol{\epsilon}, \mathbf{c}_{\text{add}}}\left[\left\|\boldsymbol{\epsilon}-\bar{\boldsymbol{\epsilon}}_{\hat{\boldsymbol{\theta}}, \boldsymbol{\phi}} (\mathbf{z},t, \mathbf{c}_{\text{text}}, \mathbf{c}_{\text{add}}) \right\|_2^2\right].
\end{align}
During the optimization process, the parameter of the pretrained model $\hat{\boldsymbol{\theta}}$ is frozen, and only the parameter ${\boldsymbol{\phi}}$ requires updating.

\subsection{Architecture Design} \label{sec-architecture}
\textbf{Head-heavy and foot-light configuration.}
BPS mainly consists of various specially designed residual blocks and zero convolution layers. The component CP, TP and EP contains 8, 2 and 12 residual blocks, respectively. Specifically, the residual block structures in CP and TP are identical and consist of two parts. The first part encompasses a convolutional structure responsible for extracting the condition feature, while the second part combines the temporal features through the scale shift norm mechanism to enable precise temporal control. In contrast, a lightweight residual structure is used in EP, where convolution plays a central role in extracting the condition feature.The detailed structural diagram is provided in Appendix \ref{app_BPS}.

\textbf{Differentiated integration structure.} \label{sec-Differentiated-integration-structure}
Fig.\ref{1d-acc-fig} and Table \ref{table-miou} provide insights of the influence by introducing additional information at different stages of the pretrained model. Interventions at earlier stages and middle blocks exhibit a more pronounced effect. With this in mind, basing on the EMD integration mode, we assign different weights to the perturbations originating from different stages. Specifically, for each scale, two residual blocks and one convolution layer are used to extract two latent features from the contracting path. These features are assigned weights $w_i = \frac{\text{scale}}{4}, 1\le i\le 8$, ensuring that earlier stage features have a greater impact. The latent features from TP are assigned a weight of $w_1$. And the perturbation terms from EP are uniformly assigned a weight of $w_i = 1, 10\le i \le 21$. To this end, the updated formulation becomes:
\begin{align}
    \hat{\mathbf{h}}^{i,j}=\mathbf{h}^{i,j}+w_i * \mathbf{v}_{\text{add}}^i, i= 1,2, ..., 21.
\end{align}

The superiority of such structure design is demonstrated in Table \ref{table-miou}.

\subsection{Task-Specific Condition Design} \label{sec-condition}

The design of conditions ($\boldsymbol{\psi}$) plays a crucial role in identifying the target domain as demonstrated in Lemma \ref{lemma}. It is imperative for BPS to work in concert with the text control layers within the SD backbone. This collaboration is necessary to harness knowledge pertaining to $\boldsymbol{\psi}$, which in turn facilitates the efficient transmutation of the information contained in $C$ into the image representation $Z_0$. Moreover, this process must effectively distinguish the signal from the noise disturbance term $Z_t$ to ensure the integrity of the target domain identification. We propose the following condition designs for several tasks.

\begin{figure}[ht]
    \vskip 0.05in
    \begin{center}
    \centerline{\includegraphics[width=\columnwidth*4/5]{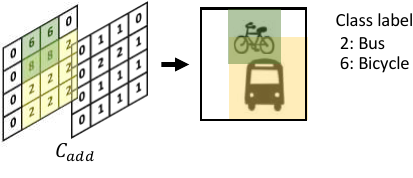}}
    \vskip -0.05in
    \caption{The process of creating a layout condition: perform addition operations on the pixel points of the first channel using layer labels, with the second channel tracking the number of operations on each bit.}
    \label{layer-condition-fig}
    \end{center}
    \vskip -0.2in
\end{figure}

\textbf{Layout condition design.} Layout-to-image generation empowers users with precise control over layers. Most existing methods contain complex layer processing and monofunctional neural network architectures \citep{cheng2023layoutdiffuse, zheng2023layoutdiffusion}. However, our BPS model effectively handles the task by converting layouts into image-type conditions with dimensions of $512 \times 512 \times 2$. As shown in Fig.\ref{layer-condition-fig}, the first channels capture object location information, while the second channel records object overlap. These conditions, combined with the proposed network structure, contribute to achieving a superior FID of 20.24 on the COCO dataset. More information regarding the experimental setup is provided in Appendix \ref{app_data}.

\begin{figure}[ht]
    \vskip 0.05in
    \begin{center}
    \renewcommand{\arraystretch}{0.5}
    \begin{tabular}{cc}
        \includegraphics[width = 0.23\columnwidth]{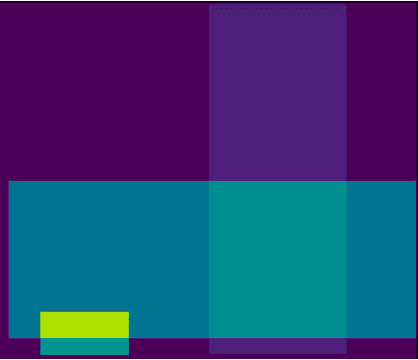} &
        \includegraphics[width = 0.23\columnwidth]{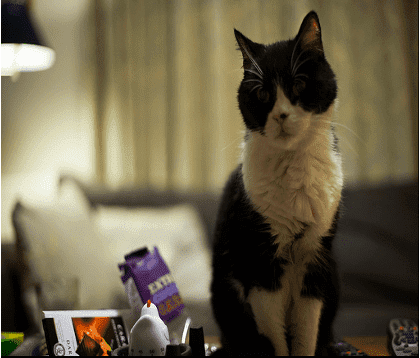}
        \includegraphics[width = 0.23\columnwidth]{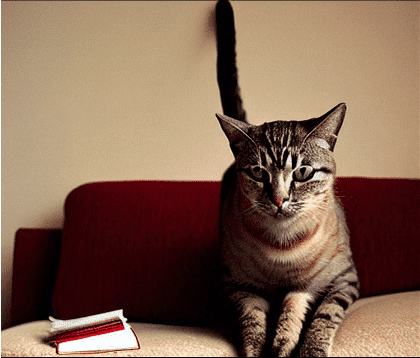}
        \includegraphics[width = 0.23\columnwidth]{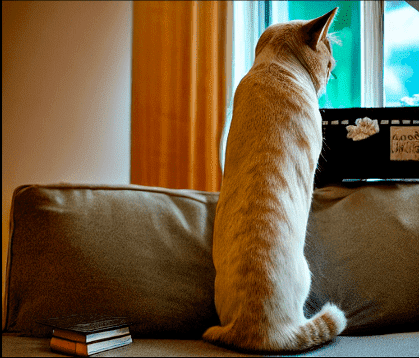} \\
        \\
        \includegraphics[width = 0.23\columnwidth, height = 0.20\columnwidth]{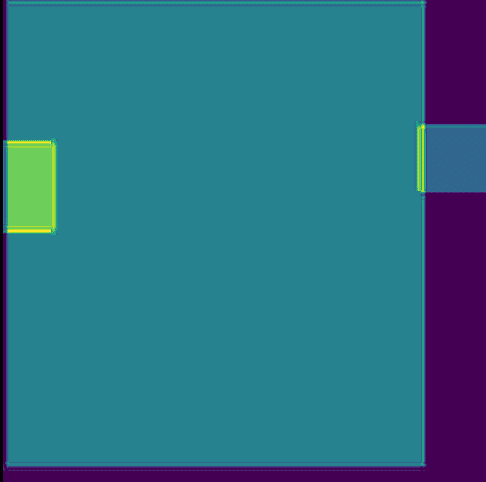} &
        \includegraphics[width = 0.23\columnwidth, height = 0.20\columnwidth]{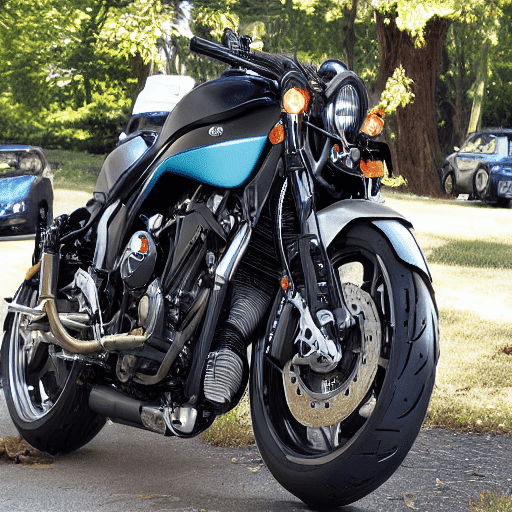}
        \includegraphics[width = 0.23\columnwidth, height = 0.20\columnwidth]{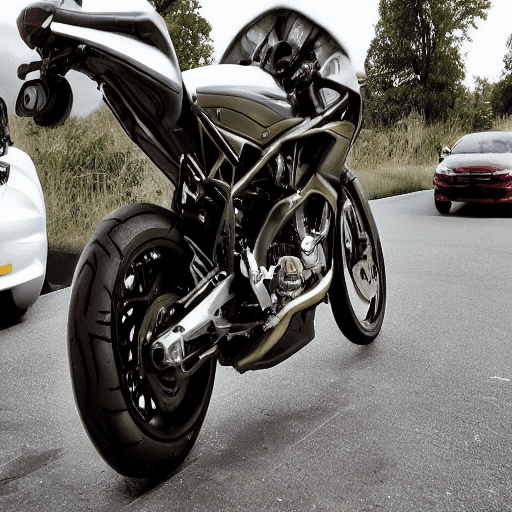}
        \includegraphics[width = 0.23\columnwidth, height = 0.20\columnwidth]{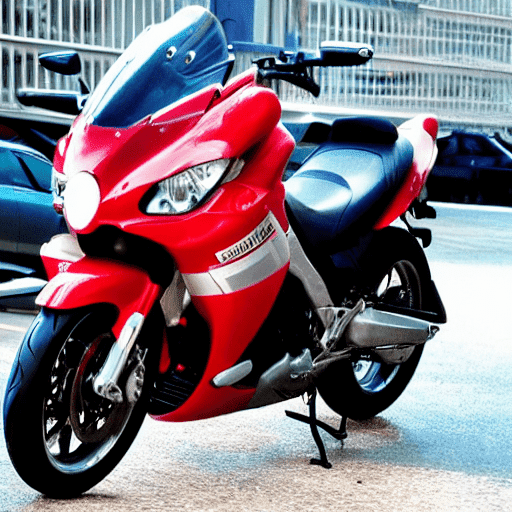}
    \end{tabular}
\end{center}
    \label{layerfig1}
    \vspace{-0.05in}
    \caption{Generated samples with layout condition.}
\end{figure}

\begin{figure}[ht]
    \begin{center}
    \centerline{\includegraphics[width=\columnwidth, height=2.2 in]{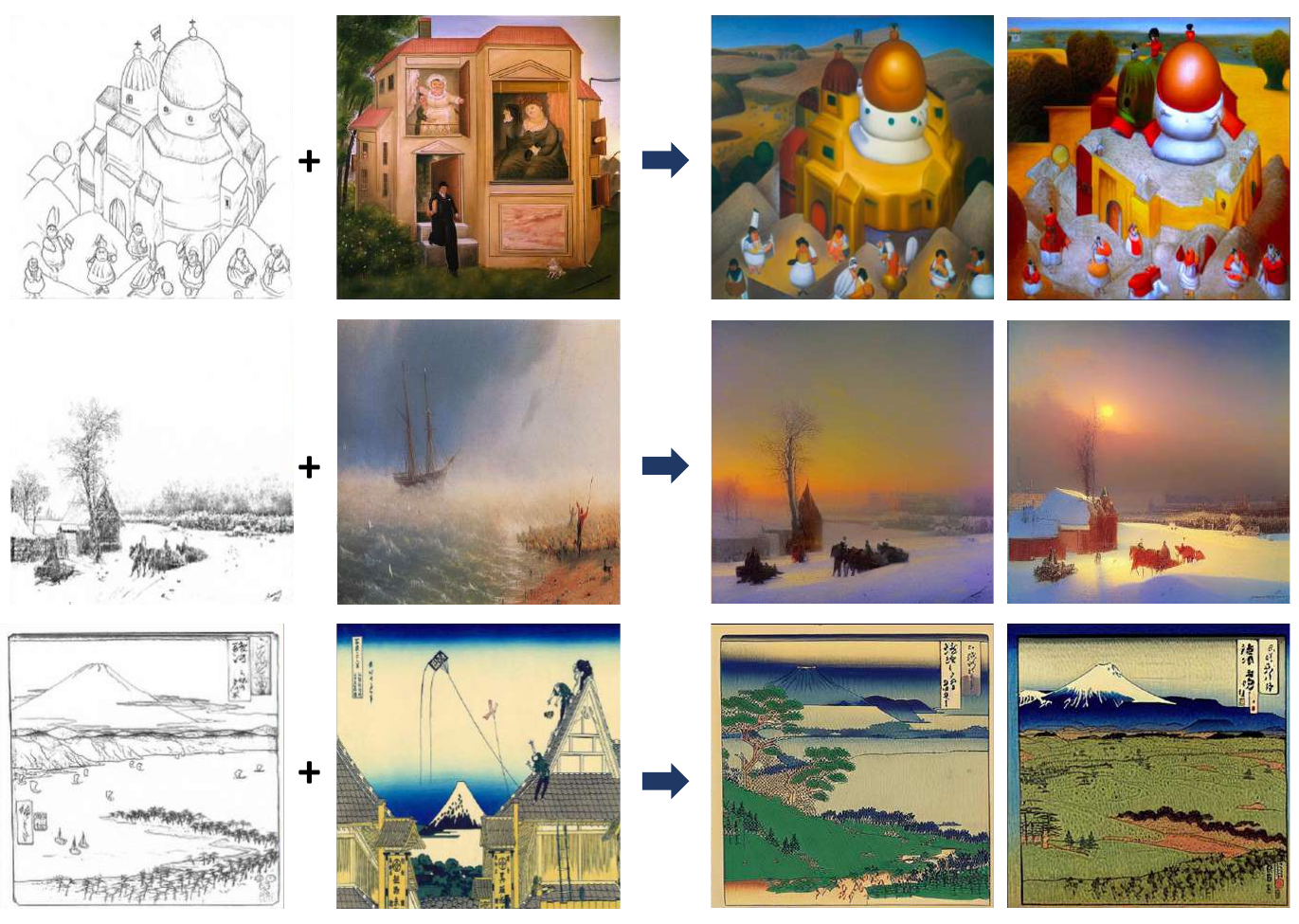}}
    \caption{Adapt the pretrained model to a specific art domain with multiple conditions (line, style).}
    \label{style-fig}
    \end{center}
    \vskip -0.2in
\end{figure}

\textbf{Multiple conditions design.} Style and line are essential elements in artistic drawing and achieving user-customized artwork. Given the abstract nature of image style, data enhancement techniques and the autoencoder of SD are employed to the latent feature extraction of the source image. The line structure is constructed using the HED soft edge method \citep{xie2015holistically}. In the end, the two conditions are concatenated as input to the BPS, enabling the generation of customized artwork, as shown in Figure \ref{style-fig}.

\begin{figure}[ht]
    \vskip 0.07in
    \begin{center}
    \centerline{\includegraphics[width=\columnwidth]{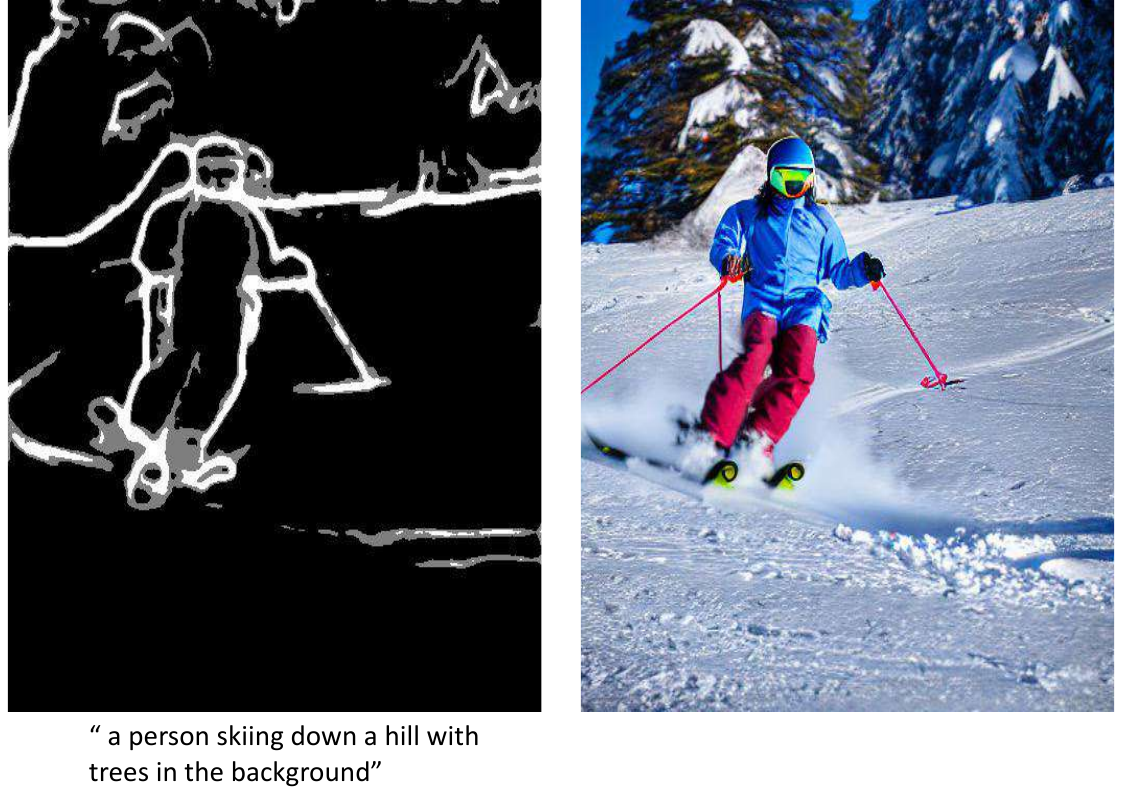}}
    \caption{ Sketch-to-image. The left showcases the sketch conditions and prompts generated by BLIP, and the right presents the corresponding outputs generated by BPS. }
    \label{sketch-fig}
    \end{center}
    \vskip -0.25in
\end{figure}

\textbf{Prompt design.} Conditions $C_{\text{text}}$ and $C_{\text{add}}$ can be highly correlated. To reduce dependence on prompts and enhance the active capture of information of $C_{\text{add}}$, we design stochastic multilevel textual control.  This approach includes three levels of text with equal probability: a generalized overview, such as the default prompt in \citet{zhang2023adding} (``a high-quality, detailed, and professional image."), followed by object descriptions (e.g., ``a car and a person"), and a detailed portrayal of the objects and their states, generated by BLIP \citep{li2022blip}. An example is presented in Fig. \ref{sketch-fig}. These levels of text are incorporated into the integration model during training, with equal probability assigned to each level.

\section{Experiments}

\subsection{Implementation Details} \label{sub_experiment}
We train multiple BPS models with diverse conditions to customize the pretrained diffusion system for various task-specific scenarios. The pretrained model used is SD V1.5. Training is performed on 2 NVIDIA A100 80GB GPUs, taking at most 48 hours to complete. The optimizer used is AdamW \citep{loshchilov2017decoupled} with a learning rate of $1 \times 10^{-5}$. The effective batch size is 256 after applying gradient accumulation.

The performance of BPS is evaluated in layout-to-Image, artistic drawing, segmentation-to-image, and segmentation-to-image tasks. Further details regarding the datasets are provided in the supplementary materials.

\subsection{Qualitative Examination}
\begin{figure}[ht]
    \begin{center}
    \subfigure{
      \includegraphics[width=0.21\columnwidth, height=0.6 in]{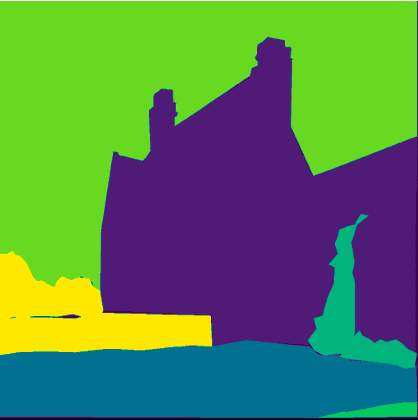}
    }
    \subfigure[Summer]{
      \includegraphics[width=0.21\columnwidth, height=0.6 in]{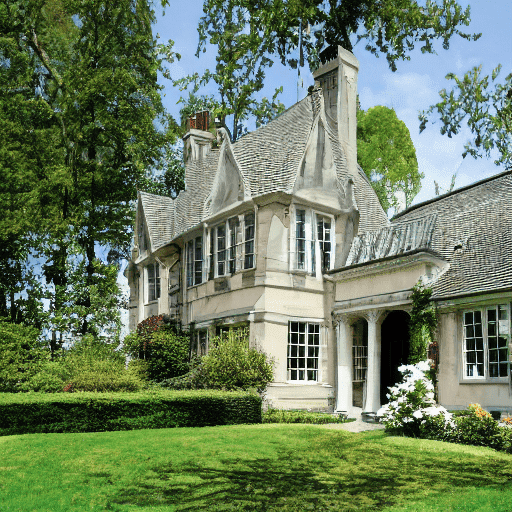}
    }
    \subfigure[Autumn]{
      \includegraphics[width=0.21\columnwidth, height=0.6 in]{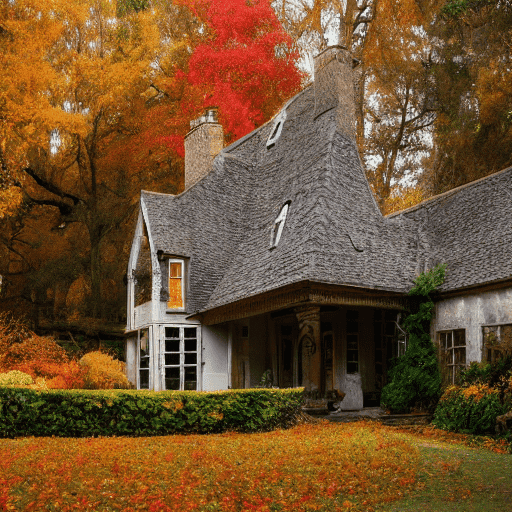}
    }
    \subfigure[Winter]{
      \includegraphics[width=0.21\columnwidth, height=0.6 in]{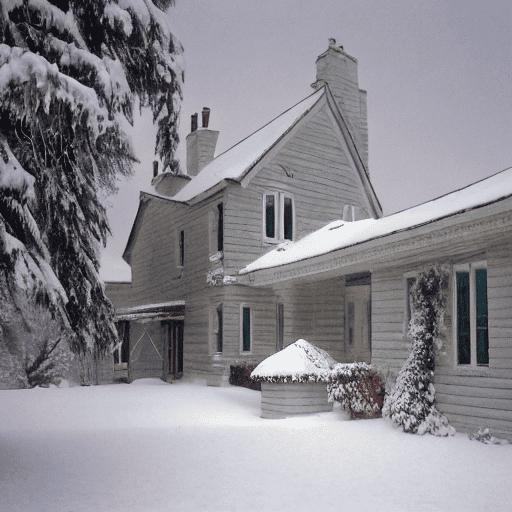}
    } \\
    \subfigure[Desert]{
      \includegraphics[width=0.21\columnwidth, height=0.6 in]{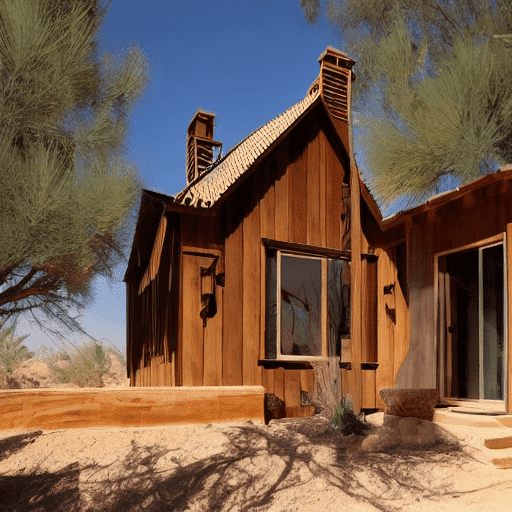}
    }
    \subfigure[Seabed]{
      \includegraphics[width=0.21\columnwidth, height=0.6 in]{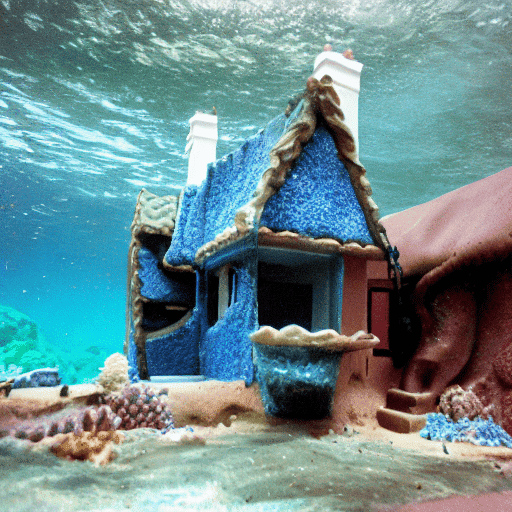}
    }
    \subfigure[Toy bricks]{
      \includegraphics[width=0.21\columnwidth, height=0.6 in]{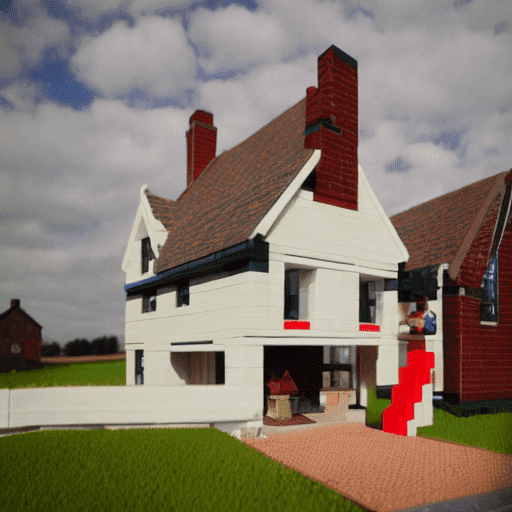}
    }
    \subfigure[Plasticine]{
      \includegraphics[width=0.21\columnwidth, height=0.6 in]{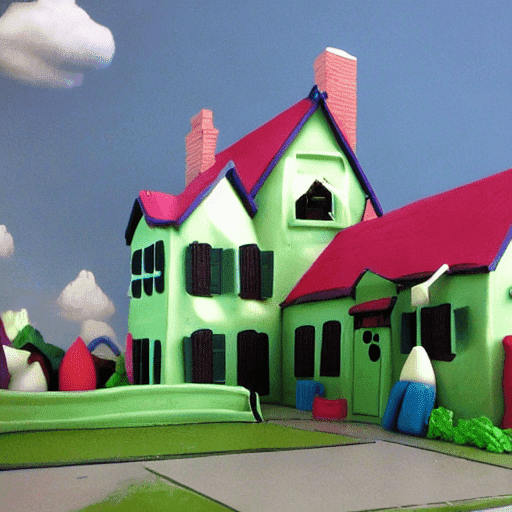}
    }
    \end{center}
    \vspace{-0.1in}
    \caption{BPS generated samples in response to prompts regarding time, place, and building materials. Refer to the supplementary material for detailed textual prompts.}
    \label{prompt-fig-2}
    \vspace{-0.1in}
\end{figure}

Figures 7, \ref{style-fig} and \ref{sketch-fig} showcase the robust performance of BPS across various tasks. Fig. \ref{prompt-fig-2} shows the generated images in several prompt settings, demonstrating its ability to accurately interpret content semantics from $C_{\text{text}}$ and incorporate both $C_{\text{add}}$ and $C_{\text{text}}$.

\subsection{Comparison}

\begin{figure*}[ht]
    \vskip 0.05in
    \begin{center}
    \centerline{\includegraphics[width=6.4 in 
    , height=4.3in
    ]{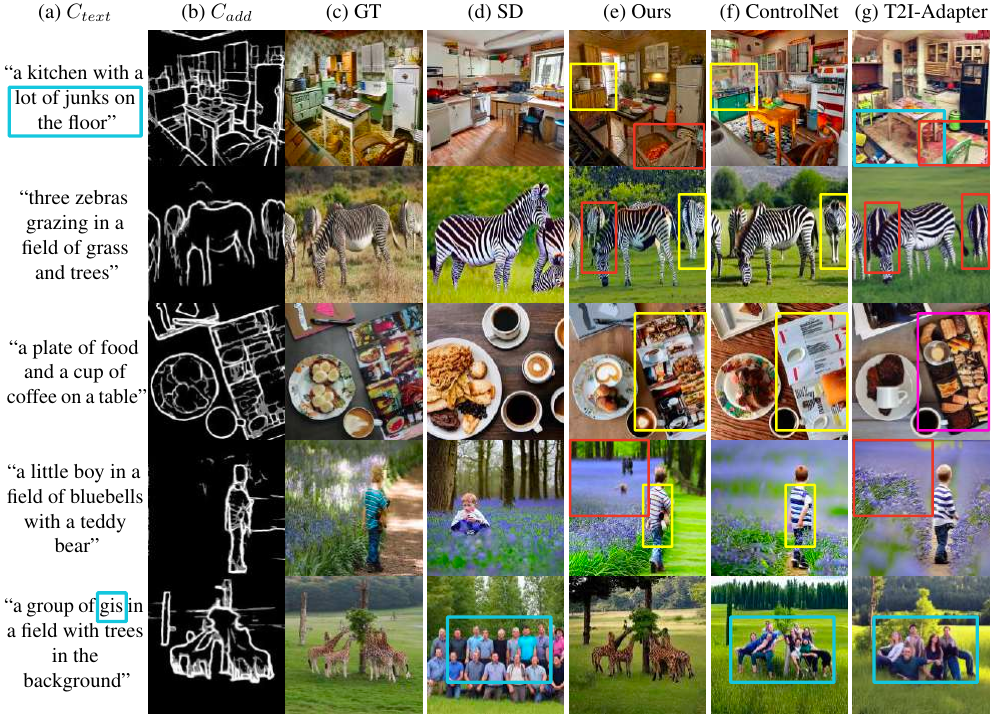}} 
    \vskip -0.05in
    \caption{Visualization comparison: our BPS, Stable Diffusion \citep{rombach2022high}, T2I-Adapter \citep{mou2023t2i}, ControlNet \citep{zhang2023adding}, and Ground Truth (GT).}
    \label{comparison-fig}
    \end{center}
    \vskip -0.27in
\end{figure*}

We compare our method with T2I-Adapter \cite{mou2023t2i}, ControlNet \cite{zhang2023adding}, and the original SD \citep{rombach2022high}, using the sketch condition in the COCO17 dataset. The dataset consists of 164K images, with the corresponding sketch maps generated by the edge prediction model proposed by \citet{su2021pixel}. T2I-Adapter, ControlNet, and BPS are trained for 10 epochs using the experimental setup in Section \ref{sub_experiment}. As shown in Fig. \ref{comparison-fig}, while T2I-Adapter exhibits imaginative capabilities (as indicated by the pink box), both T2I-Adapter and ControlNet exhibit limited attention to details, as evident in the red and yellow boxes. In contrast, BPS demonstrates a refined and hierarchical treatment of details. Furthermore, even in the presence of significant textual errors, BPS demonstrates the ability to achieve relatively unaffected generation quality. However, SD, ControlNet, and T2I-Adapter are noticeably impacted, as highlighted in the blue box.

\begin{table}[ht] 
    \vskip -0.07in
    \caption{ User Preference Ranking for Result Quality and Condition Fidelity. Score 1 to 3 indicates worst to best.}
    \label{table-human}
    \vskip 0.15in
    \begin{center}
    \begin{small}
    \begin{tabular}{lccr}
    \toprule
     & T2I-Adapter & ControlNet & BPS \\
    \midrule
    Result Quality & $1.868$ & $1.77$ & $\mathbf{2.38}$  \\
    \midrule
    Condition Fidelity & $1.95$ & $1.53$ & $\mathbf{2.52}$  \\
    \bottomrule
    \end{tabular}
    \end{small}
    \end{center}
    \vskip -0.05in
\end{table}

\begin{table*}[htb] 
    \caption{Quantitative comparison (FID/CLIP Score). Stable diffusion employs text as the condition, while the other methods utilize text+sketch as the condition.}
    \label{table-fid}
    \vskip 0.15in
    \begin{center}
    \begin{small}
    \begin{tabular}{lcccccr}
    \toprule
     & SD & T2I-Adapter (Epoch=10) &ControlNet & BPS (Epoch=10) & BPS (Epoch=15) \\
    \midrule
    FID $\downarrow$ & $20.59$& $18.39$ & $19.41$  & $10.49$ & $\mathbf{10.04}$\\
    \midrule
    CLIP Score $\uparrow$ & $\mathbf{0.2647}$ &$0.2642$& $0.2361$  & $0.2614$& $0.2614$ \\
    \bottomrule
    \end{tabular}
    \end{small}
    \end{center}
    \vskip -0.15in
\end{table*}

For quantitative evaluation, we employ FID \citep{Seitzer2020FID}, CLIP Score (ViT-L/14, \citet{radford2021learning}), as presented in Tab.\ref{table-fid}. We randomly sample 5000, 2500, and 2500 images from the validation set, training set, and testing set, respectively, to obtain the conditions for generation. While there is a slight difference in the CLIP Score due to the impact of the stochastic multilevel textual control we introduced, BPS achieves the best FID score. Furthermore, we conduct the human evaluation following the experimental design in \citet{zhang2023adding}. We sample 25 unseen hand-drawn sketches and assign each sketch to three methods: T2I-Adapter sketch model, ControlNet, and BPS. We invite 20 users to rank these 25 groups of three results based on ``the image quality showcased" and ``the fidelity to the sketch." User Preference Ranking is used as the preference metric, where users rank each result on a scale of 1 to 3 (lower is worse). As shown in Tab. \ref{table-human}, the results indicate a preference for BPS among the users.

\subsection{More Investigation}
Additionally, we examine the noteworthy considerations in applications, i.e. time efficiency, data scarcity, and intervention strength.

\begin{figure}[ht]
    \vskip 0.07in
   \begin{center}
   \centerline{\includegraphics[width=0.9\columnwidth]{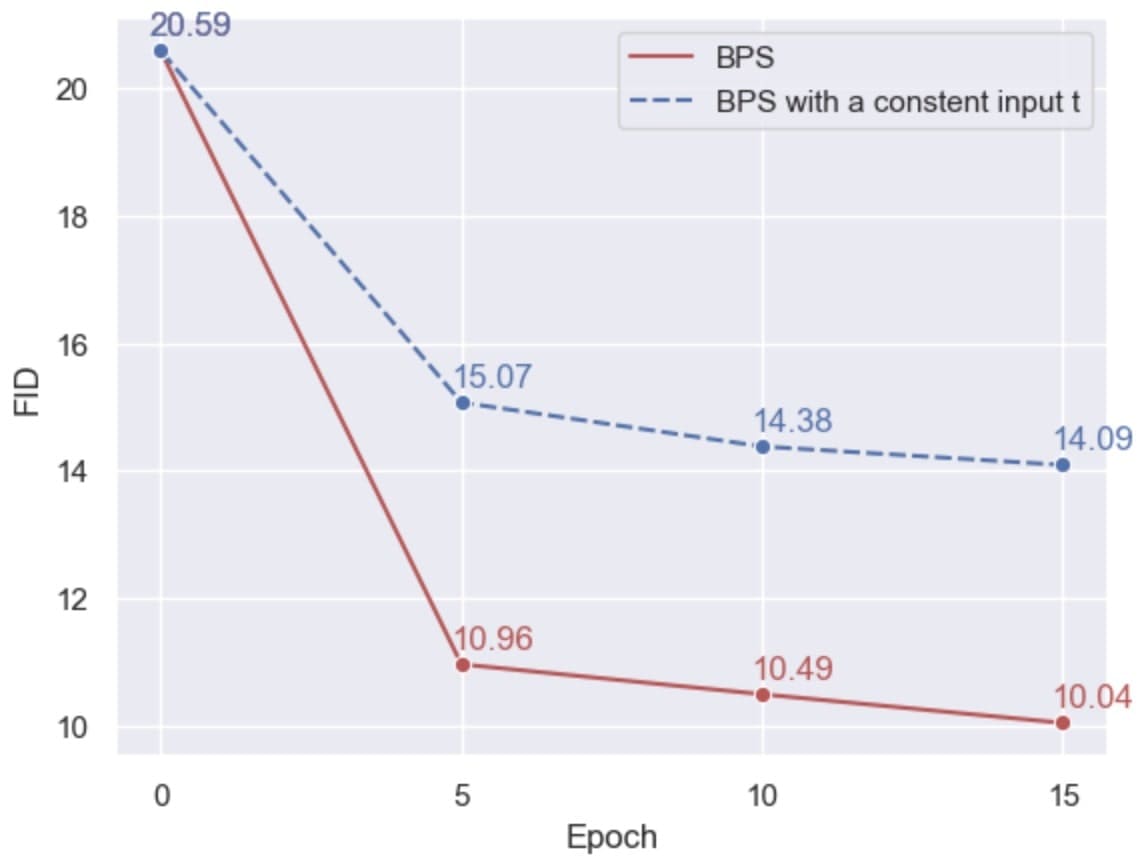}}
   \vskip -0.05in
   \caption{Performance comparison of BPS with constant time input and the original BPS.}
   \label{fig-time}
   \end{center}
   \vskip -0.3in
\end{figure}

\textbf{Implication of time-step information.} In the sketch-to-image task, we conduct a controlled experiment by setting the time input of BPS to a constant value, and follow the other setup outlined in Section \ref{sub_experiment}. According to Fig.\ref{fig-time}, the performance declines throughout various training stages when time-step information is absent. It highlights the need to incorporate an additional input, $t$, into the learning module to enable adaptive adjustment of intervention intensity in the original gradient direction.

\textbf{Time efficiency.} We study time efficiency by training the model up to 15 epochs using the COCO17 dataset. Fig.\ref{fig-time} illustrates the FID score of the integrated SD with BPS. It demonstrates satisfactory performance as early as the fourth iteration of training. Moreover, comparing to the initial point, BPS significantly improves the performance of SD.

\begin{figure}[ht]
    \vskip 0.05in
   \begin{center}
   \centerline{\includegraphics[width=\columnwidth]{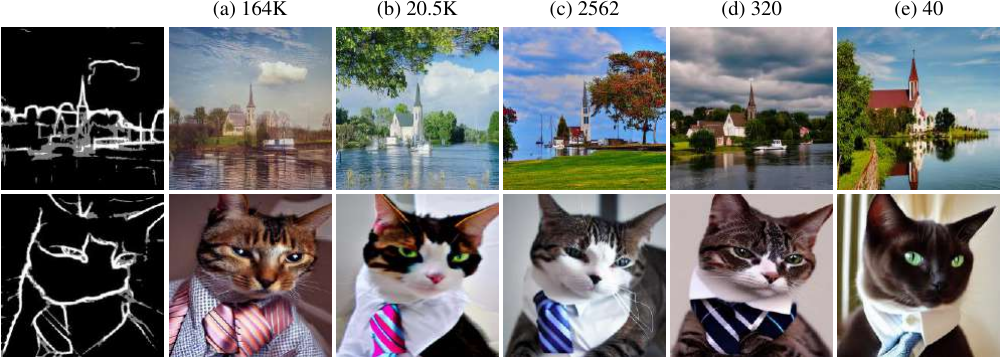}}
   \caption{The generation results of models trained with different training set sizes.}
   \label{abla-n-fig}
   \end{center}
   \vskip -0.2in
\end{figure}

\textbf{Data scarcity.} To assess the model's generalization ability in scenarios with limited data, we randomly select subsets of 40, 320, 2562, 20500, and 164K images from the training and validation sets for fine-tuning. Fig.\ref{abla-n-fig} presents the fine-tuning performance after 15 iterations using sketches from the test dataset. Our model demonstrates exceptional performance even with a limited number of samples. More results are in the Appendix.

\begin{figure}[ht]
      \vskip 0.1in
       \begin{center}
   \centerline{\includegraphics[width=0.9\columnwidth]{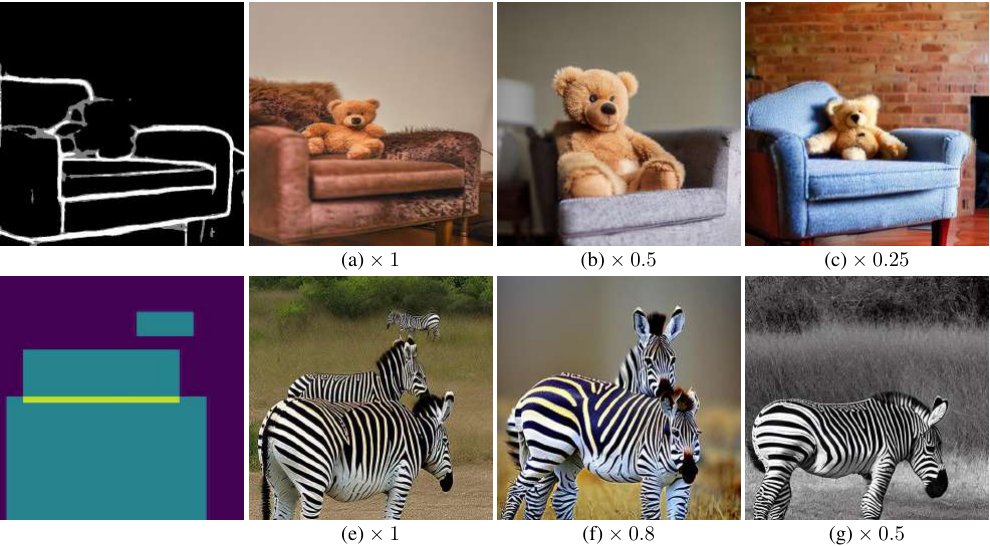}}
       \caption{
       The impact of intervention weights on generated results in layout-to-image and sketch-to-image tasks. The subheading indicates the multiplier applied to the initial weight.}
       \label{weight-fig}
       \end{center}
       \vskip -0.3in
   \end{figure}

\textbf{Intervention ability.} The findings depicted in Fig.\ref{weight-fig} underscore the significant influence of the intervention weights ($w_i$, where $1 \le i \le 21$) within the BPS framework on the outcomes of the interventions. Specifically, decreasing these weights contributes to an increase in diversity among the generated results.

\section{Discussion and Limitations}

In this paper, we formulate the problem of generative model domain adaptation and propose a Bayesian fine-tuning framework, which uses the score functions of a pretrained  diffusion model as the \textit{prior scores} and updates them to obtain the \textit{posterior scores} using  Bayes' theorem. It explores two crucial aspects of domain adaptation: the choice of approximation functions, represented by the neural network structure, and the definition of conditions specific to the task domain. To tackle the first aspect, we introduce a neural network architecture,  Bayesian power steering, for the implementation of the Bayesian formulation in hierarchical
levels of the feature space. Our BPS offers several  advantages, 
including  (1) exceptional domain recognition and control across different tasks, (2) a compact parameterization and fast convergence, and (3) suitability for data-scarce scenarios with broad applicability. However, our approach still faces certain challenges that warrant further investigation: (a) developing a more refined structure to capture the degree of intervention for multiple conditional inputs, and (b) adaptive adjustment of intervention weights to achieve desired outcomes in extended domain scenarios. These issues remain areas for future study.

\section*{Acknowledgment}
We extend our sincere appreciation to the Area Chair and the four anonymous reviewers for their time, expertise, and insightful comments. This research was conducted using the computing resources provided by the Research Center for the Mathematical Foundations of Generative AI in the Department of Applied Mathematics at The Hong Kong Polytechnic University. Ting Li's research is partially supported by the National Natural Science Foundation of China (Grant No.12301360) and the research grants from The Hong Kong Polytechnic University. Jian Huang's work is supported by the research grants from The Hong Kong Polytechnic University.

\section*{Impact Statement}

This paper presents work whose goal is to advance the field of Machine Learning. There are many potential societal consequences of our work, none which we feel must be specifically highlighted here.

\newpage

\nocite{langley00}

\bibliography{BPS_bib}
\bibliographystyle{icml2024}

\newpage
\appendix
\onecolumn
\begin{center}
    \textbf{\large APPENDIX}
\end{center}

In the appendix, we provide supplementary theoretical derivations for the BPS introduced in Section 3. Furthermore, we include more detailed description of the implementation of BPS and additional experimental results.




\section{Derivation of Equation (\ref{s3_idea})} \label{app_derivation}
Recall that we have
$$
\boldsymbol{\epsilon}^*\left(\mathbf{z}_t, t, \mathbf{c}_{\mathrm{text}}\right):=\mathbb{E}\left[\boldsymbol{\eta} \mid \mathbf{z}_t, \mathbf{c}_{\mathrm{text}}\right]=-\sqrt{1-\bar{\alpha}_t} \nabla \log p\left(\mathbf{z}_t \mid \mathbf{c}_{\mathrm{text}}\right)
$$ through Tweedie's formula \citep{efron2011tweedie}, and
\begin{align} \label{equation1}
    p(\mathbf{z}_t \mid \mathbf{c}_{\text {text }}, \mathbf{c}_{\mathrm{add}}) =\frac{p(\mathbf{c}_{\mathrm{add}} \mid \mathbf{c}_{\text {text }}, \mathbf{z}_t) p(\mathbf{z}_t \mid \mathbf{c}_{\text {text }})}{p(\mathbf{c}_{\mathrm{add}} \mid \mathbf{c}_{\text {text }})}
\end{align} by Bayes' theorem.

Take the logarithm of both sides of Equation (\ref{equation1}) and subsequently compute the gradient with respect to $\mathbf{z}_t$, then we obtain
\begin{align*}
     \nabla \log p (\mathbf{z}_t \mid \mathbf{c}_{\mathrm{text}}, \mathbf{c}_{\mathrm{add}}) &= \nabla \log p(\mathbf{c}_{\mathrm{add}} \mid \mathbf{c}_{\mathrm{text}}, \mathbf{z}_t) +\nabla \log p(\mathbf{z}_t \mid \mathbf{c}_{\mathrm{text}}) -\nabla \log p(\mathbf{c}_{\text {add }} \mid \mathbf{c}_{\text {text }}) \\
& =\nabla \log p(\mathbf{c}_{\text {add }} \mid\mathbf{c}_{\text {text }}, \mathbf{z}_t) +\nabla \log p(\mathbf{z}_t \mid \mathbf{c}_{\mathrm{text}}),
\end{align*} where $p(\mathbf{c}_{\text {add }} \mid\mathbf{c}_{\mathrm{text}})$ is constant for $\mathbf{z}_t$ so that $\nabla \log p(\mathbf{c}_{\text {add }} \mid \mathbf{c}_{\text {text }})=0$.

Combining the above formulas, the integrated denoise function can be derived as follows.
\begin{align*}
 \overline{\boldsymbol{\epsilon}}^*(\mathbf{z}_t, t, \mathbf{c}_{\mathrm{text}}, \mathbf{c}_{\mathrm{add}}) &:=\mathbb{E}\left[\boldsymbol{\eta} \mid \mathbf{z}
_t, \mathbf{c}_{\mathrm{text}}, \mathbf{c}_{\mathrm{add}}\right] =-\sqrt{1-\bar{\alpha}_t} \nabla \log p(\mathbf{z}_t \mid \mathbf{c}_{\mathrm{text}}, \mathbf{c}_{\mathrm{add}}) \\
&=\quad-\sqrt{1-\bar{\alpha}_t}  \left[\nabla \log p(\mathbf{c}_{\mathrm{add}} \mid \mathbf{z}_t, \mathbf{c}_{\mathrm{text}}) +\nabla \log p(\mathbf{z}_t \mid \mathbf{c}_{\mathrm{text}})\right] \\
&=\quad-\sqrt{1-\bar{\alpha}_t} \nabla \log p(\mathbf{c}_{\mathrm{add}} \mid \mathbf{z}_t, \mathbf{c}_{\mathrm{text}}) +\boldsymbol{\epsilon}^*(\mathbf{z}_t, t, \mathbf{c}_{\text{text}}),
\end{align*} which complete the derivation.

\begin{figure}[ht]
    \begin{center}
    \centerline{\includegraphics[width=\textwidth]{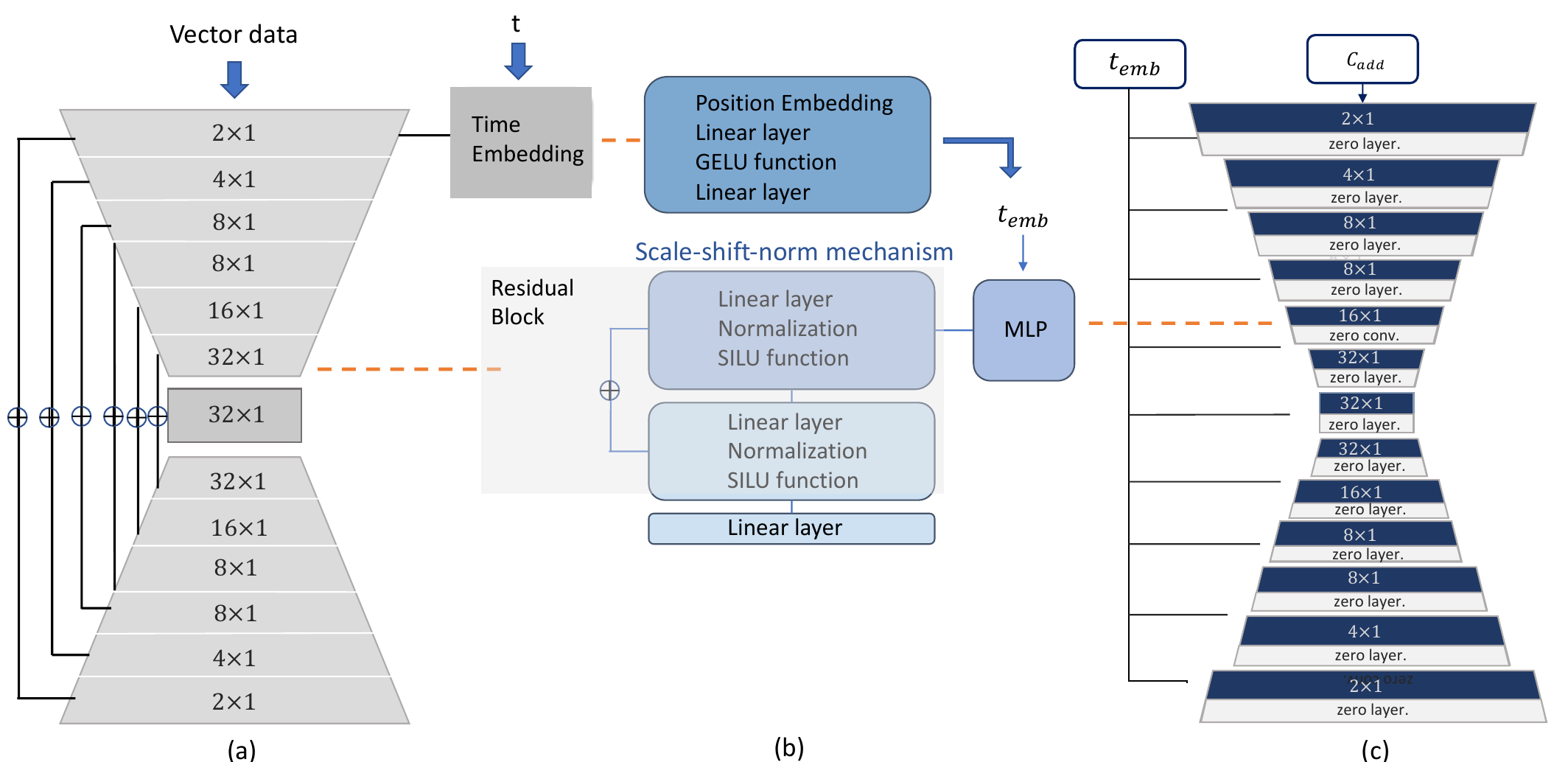}}
    \caption{Architecture of pre-trained model and learnable modules for vector data.}
    \label{1dnet-supp-fig}
    \end{center}
\end{figure}

\section{Generation of 2D Data}
In this setting, the samples exhibit a multimodal distribution, with a probability of $0.7$ uniformly distributed on the first ring and $0.3$ on the second ring. The centers of the rings are positioned at coordinates $(0, 0)$ and $(5, 0)$ in the 2D coordinate space. Both rings have identical sizes and feature inner and outer radii of $0.6$ and $1$ respectively.

During the pretraining stage, we randomly sampled 500K samples from the rings to train the prior denoise model. In the fine-tuning stage, only 100 samples with support set information (i.e., labels indicating the originating ring) were available for training. The objective is to generate samples adhering to the specified ring distribution.

\subsection{Network Structure}
We develop a U-net architecture specifically designed for vector data. This pre-trained U-Net model, shown in Fig. \ref{1dnet-supp-fig}(a), serves as the backbone for fine-tuning. Data features are scaled into 2, 4, 8, 16, and 32 dimensions across blocks, each comprising a Residual block and a linear layer, as depicted in Fig. \ref{1dnet-supp-fig}(b).

The fine-tuning process involves learnable modules following the structure illustrated in Fig. \ref{1dnet-supp-fig}(c). Notably, these modules incorporate a ``zero layer" - a fully connected layer initialized with zero weights. The intermediate outputs from the learnable module are first processed through the zero layer before being combined with the outputs of the pre-trained model. This ensures a smooth evolution of the original signals (i.e., the outputs of the pre-trained model) during the training process, facilitating model convergence.

\subsection{Quantitative Evaluation}
We evaluate distinct integration modes to fine-tune the pre-trained model. After generating samples using the ground truth labels and the integrated model, we employ a linear binary classifier to determine the model to which the samples belong. The computed accuracy and precision metrics are then used to assess the efficacy of the integrated model configurations.

\begin{figure}[H]
    \begin{center}
    \centerline{\includegraphics[width=0.9\textwidth]{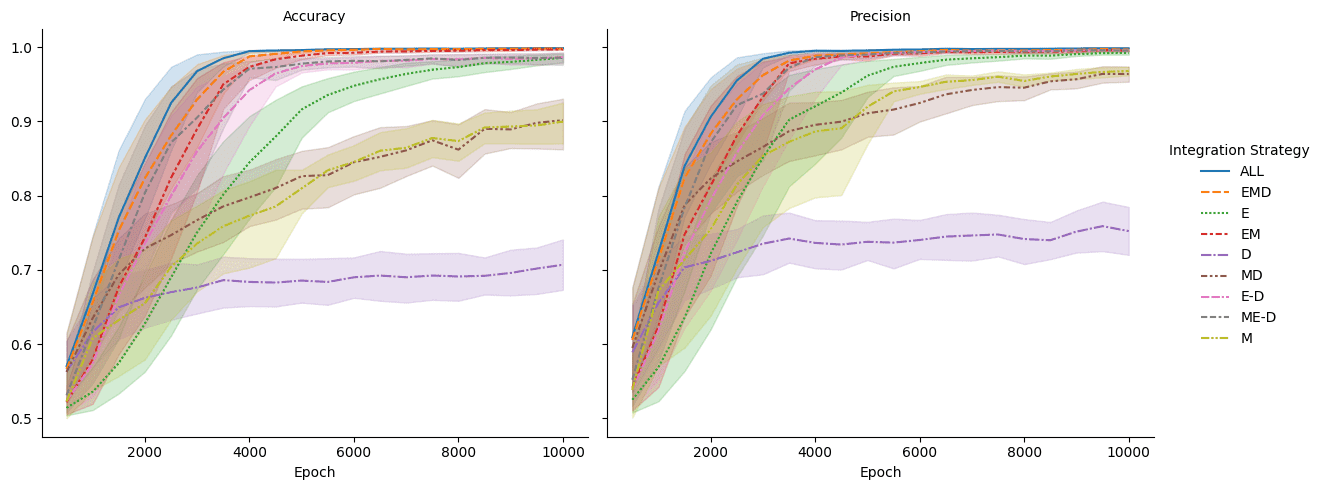}}
    \caption{Accuracy and precision of model classification using generated samples from aforementioned integration modes at different training stages with fine-tuning sample size $n=12$. Five groups of samples are randomly selected for experimentation across different modes.}
    \label{1d-acc-pre-fig}
    \end{center}
    \vskip -0.35in
\end{figure}

As shown in Fig. \ref{1d-acc-pre-fig},
the integrated modes ALL, EMD, and EM achieve the best performance in terms of both metrics and exhibit the fastest convergence rates. Although extended training allows modes ME-D, E-D, and E to approach similar precision levels, they still trail the performance of modes ALL, EMD, and EM in terms of accuracy, even with prolonged training.  In contrast, the models under modes M, MD, and D exhibit the poorest overall performance. These results preliminarily suggest that interventions on the early intermediate outputs of the pre-trained model are crucial for optimal fine-tuning.

Furthermore, to model realistic scenarios with limited data resources, fine-tuning is performed using a small number of labeled samples, ranging from 12 to 400. In this experiment, we use accuracy as the evaluation metric to validate the performance of the different integrated models. This approach allows us to assess the robustness of these integration models under resource-constrained settings.

\begin{figure}[H]
    \begin{center}
    \centerline{\includegraphics[width=\textwidth]{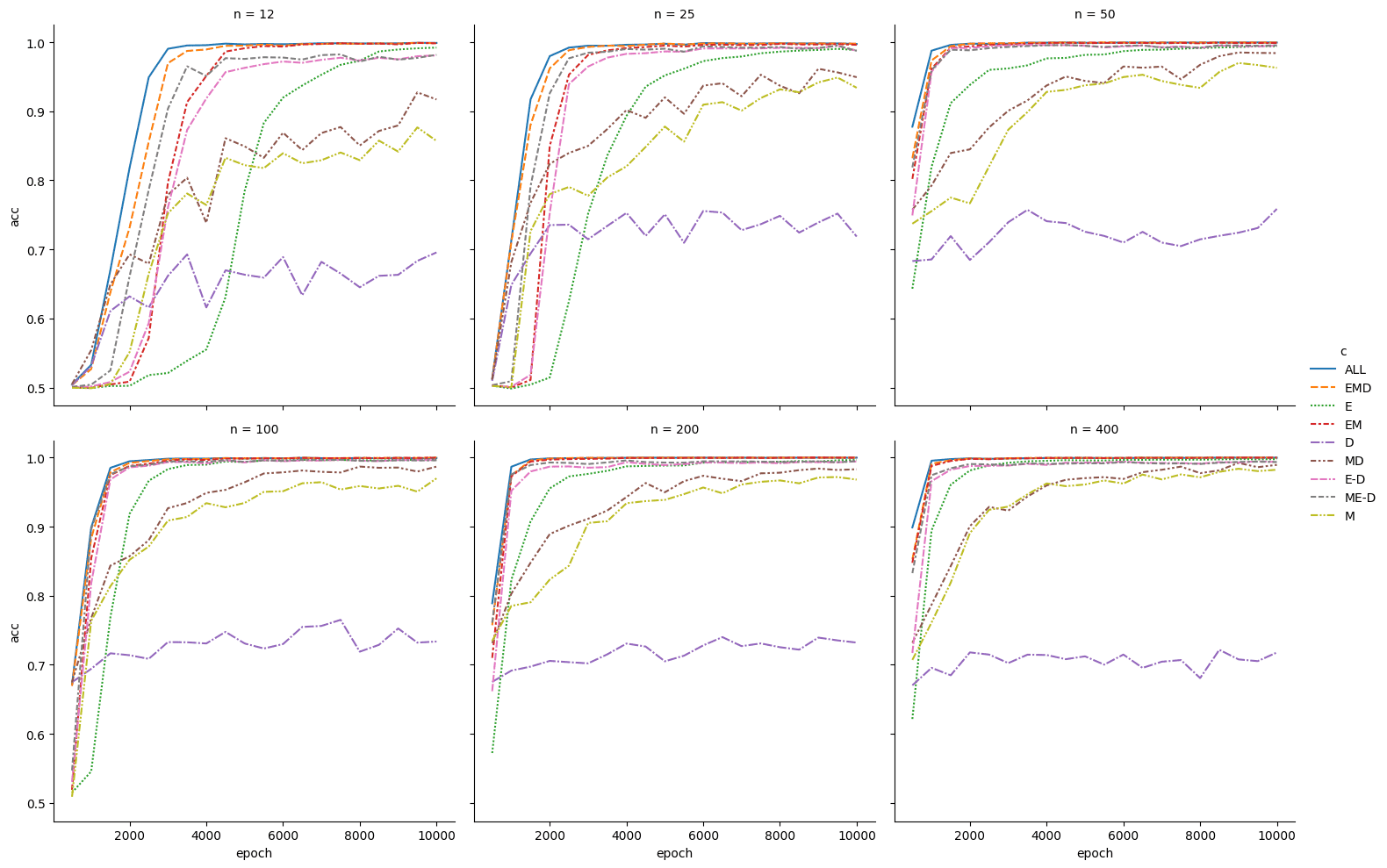}}
    \caption{Accuracy of model classification using generated samples from distinct integration modes across data sample sizes.}
    \label{1d-nn-fig}
    \end{center}
    \vskip -0.35in
\end{figure}

Notably, even as the training sample size decreases, the integrated modes ALL, EMD, and EM exhibit the highest robustness, with their performance accuracy remaining relatively stable at the peak level. In contrast, the other modes experience varying degrees of decline in convergence speed and overall performance, with models under modes M and M-D showing the most significant drops. And modes ME-D and E-D demonstrate relative instability under the most data-constrained conditions (n=12, 25). This trend is more clearly illustrated in the comparison of different models at the same training epoch, as shown in the figure below.

\begin{figure}[H]
    \begin{center}
    \centerline{\includegraphics[width=\textwidth]{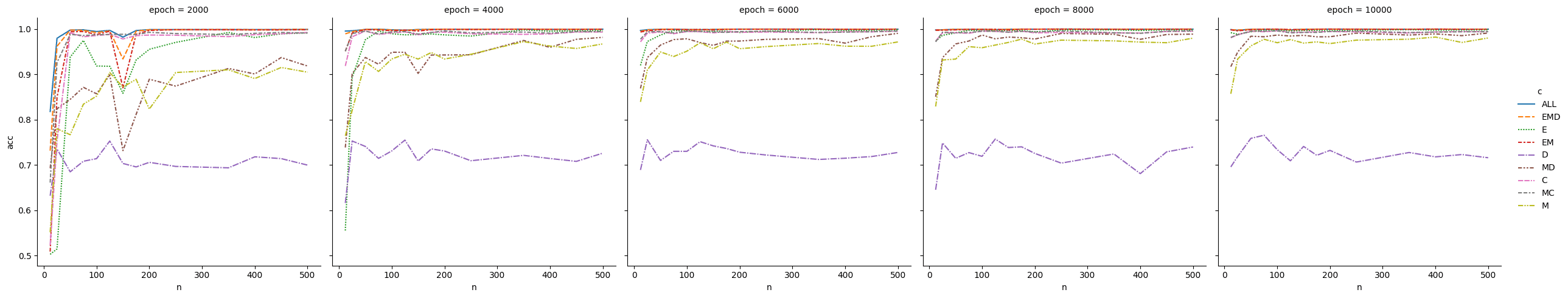}}
    \caption{Accuracy of model classification using generated samples from distinct integration modes across training stages.}
    \label{1d-epoch-fig}
    \end{center}
\end{figure}

\section{Image Generation}

\subsection{Training Parameters} \label{app_data}

We train multiple BPS models for various tasks, as detailed below.

\textbf{Segmentation-to-image}. For this structure condition, we utilize ADE20K \citep{zhou2017scene}, consisting of 25,574 images, as the training dataset. Its semantic segmentation encompasses a remarkable 193,238 annotated object parts, including intricate sub-parts. The BPS model is trained using 2 NVIDIA A100 80GB GPUs, with the Stable Diffusion V1.5 as the base model. The batch size is set to 256, achieved through a physical batch size of 16 and 4$\times$ gradient accumulation. We do not use ema weights.


\textbf{Layout-to-Image}. We evaluate the performance of bounding box layout-to-image using the COCO17 dataset \citep{lin2014microsoft}. Data preprocessing follows the approach outlined in \citet{cheng2023layoutdiffuse}, where we filter images to contain a range of 3 to 8 objects. Additionally, objects occupying less than $2\%$ of the image area are excluded. Consequently, our training dataset comprises 112,680 samples, and the validation dataset contains 3,097 samples.
The BPS model is trained using 2 NVIDIA A100 80GB GPUs, with the Stable Diffusion V1.5 as the base model. The effective batch size is 128 after applying gradient accumulation. We do not use ema weights.

\textbf{Artistic drawing}. To train our model for this application, we utilize the WikiArt Dataset \citep{tan2018improved}. The dataset was carefully selected to include works from painters with more consistent styles, resulting in a training set of 6,285 samples from 17 artists and 12 styles. The artists included in the dataset are Camille Pissarro, Childe Hassam, Claude Monet, Edgar Degas, Eugene Boudin, Gustave Dore, Ivan Aivazovsky, Marc Chagall, Pablo Picasso, Paul Cezanne, Pierre Auguste Renoir, Raphael Kirchner, Fernando Botero, Bernardo Bellotto, Cornelis Springer, Katsushika Hokusai, and Hiroshige. The art styles represented in the dataset span various genres, including Impressionism, Realism, Pointillism, Romanticism, Naive Art Primitivism, Cubism, Analytical Cubism, Synthetic Cubism, Post Impressionism, Art Nouveau Modern, Rococo, and Ukiyo-e. We train the BPS model on 2 NVIDIA A100 80GB GPUs. The batch size is set to 512, achieved through a physical batch size of 32 and 16$\times$ gradient accumulation. We do not use ema weights.

\textbf{Sketch-to-image}. For this application, we leverage the COCO17 dataset \citep{lin2014microsoft} as our training data, comprising a rich collection of 164,000 images. To generate the corresponding sketch maps, we employ the edge prediction model proposed by  \cite{su2021pixel}.  The BPS model is trained using 4 NVIDIA A100 80GB GPUs, with the Stable Diffusion V1.5 as the base model. The batch size is set to 256, achieved through a physical batch size of 32 and 8$\times$ gradient accumulation. We do not use ema weights.

\begin{figure}[H]
    \vskip -0.1in
    \begin{center}
    \centerline{\includegraphics[width=0.7\textwidth]{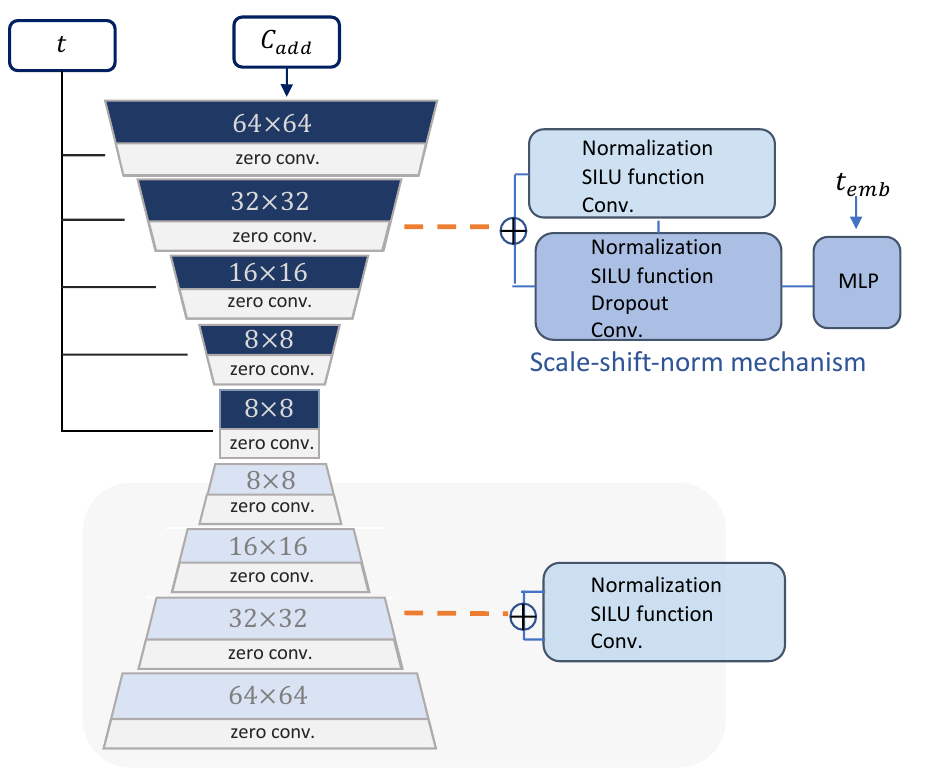}}
    \vskip -0.1in
    \caption{Architecture of BPS.}
    \label{BPS-supp-fig}
    \end{center}
    \vskip -0.35in
\end{figure}

\subsection{Network Structure: BPS} \label{app_BPS}
The BPS network structure employs a top-heavy design, as depicted in Fig \ref{BPS-supp-fig}. The dark block incorporates a residual block that primarily utilizes a Scale-shift-norm mechanism to fuse temporal and conditional information. On the other hand, the light block consists of a layer of residual blocks. Notably, the blocks incorporate several ``zero conv." - convolution layers initialized with zero weights. The intermediate outputs from the learnable module are first processed through the zero convolution layer before being combined with the outputs of the pre-trained model. This ensures a smooth evolution of the original signals (i.e., the outputs of the pre-trained model) during the training process, facilitating model convergence \citep{zhang2023adding}.

\subsection{Inference Settings}

\textbf{Sampler}: we employ Denoising Diffusion Implicit Model (DDIM) as the sampler for our approach. We utilize 50 steps to sample each image.

\textbf{Prompts}: the segmentation-to-image and Layout-to-Image tasks utilize object descriptions as text prompts (e.g., ``oven, microwave, book"). For the Artistic Drawing task, the description includes both the subject matter and the artist (e.g., ``chestnut trees at Louveciennes from Camille Pissarro's brush"). Lastly, the Sketch-to-image task adopts prompts generated by the BLIP method \citep{li2022blip}.

\subsection{Quantitative evaluation}

\textbf{Segmentation-to-image}. The performance of the BPS model is evaluated on the ADE20K dataset \citep{zhou2017scene} using the Mean Intersection-over-Union (mIoU) metric to assess conditioning fidelity. The state-of-the-art segmentation method, OneFormer  \citep{jain2023oneformer}, achieves an mIoU of 0.58 on this dataset. We generate images using segmentations from the ADE20K validation set and then feed the generative results to OneFormer for segmentation detection and computation of reconstructed IoUs. The BPS model achieve a score of 0.366, outperforming the ControlNet approach \citep{zhang2023adding}, which reported a score of 0.35.

\textbf{Layout-to-Image}. The generated images are evaluated at a resolution of $256 \times 256$. We adopt the Fréchet Inception Distance (FID) metric, as used in \cite{cheng2023layoutdiffuse}, to assess the performance of the BPS model on this task.  After training for 30 and 60 epochs, the BPS model achieves FID scores of $20.47$ and $20.24$, respectively, which are comparable to the state-of-the-art score of $20.27$ reported in \cite{cheng2023layoutdiffuse}. This indicates that the BPS model, even with a relatively simple network structure, can achieve results on par with the current state-of-the-art.

\textbf{Sketch-to-image}. We employ FID \citep{Seitzer2020FID}, CLIP Score (ViT-L/14, \citet{radford2021learning}), as presented in Tab. \ref{table-fid}. We randomly sample 5000, 2500, and 2500 images from the validation set, training set, and testing set, respectively, to obtain the sketch conditions for generation. While there is a slight difference in the CLIP Score due to the impact of the stochastic multilevel textual control we introduced, BPS achieve the best FID score, surpassing the competitio.

\subsection{Qualitative Evaluation }

\begin{figure}[H]
    \begin{center}
    \subfigure[$C_{\text{add}}$]{
      \includegraphics[width=0.21\textwidth]{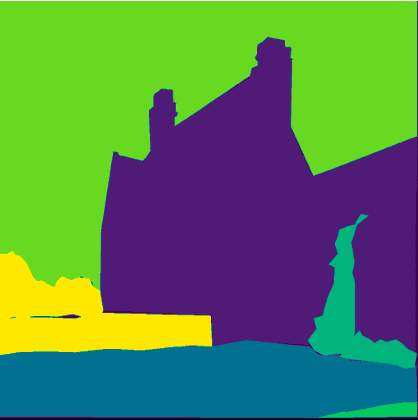}
    }
    \hspace*{\fill}
    \subfigure[``Nestled amidst lush foliage, the country house unveils its enchanting beauty as summer cascades upon it."]{
      \includegraphics[width=0.21\textwidth]{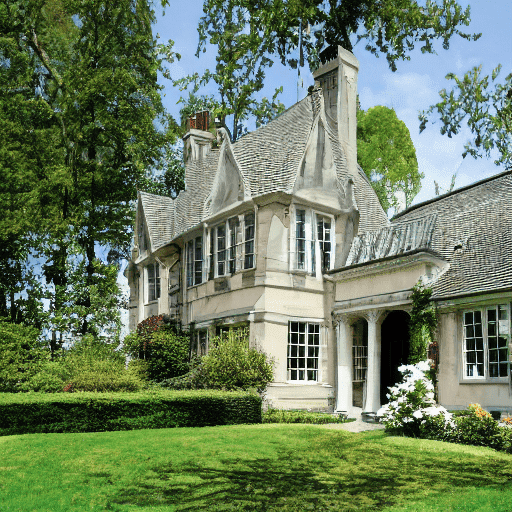}
    }
    \hspace*{\fill}
    \subfigure[``Amidst the fall's magical embrace, a country house stands adorned with a vibrant carpet of fallen leaves."]{
      \includegraphics[width=0.21\textwidth]{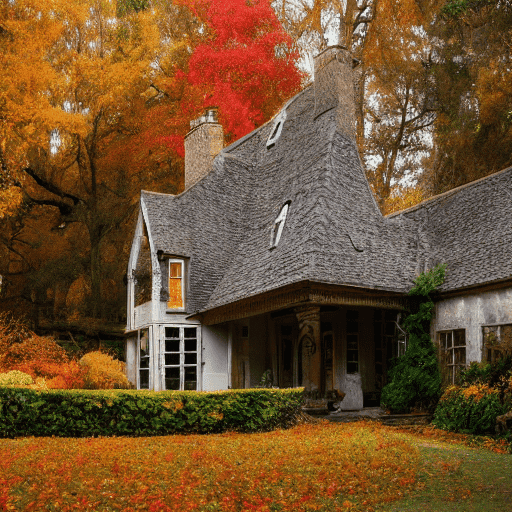}
    }
    \hspace*{\fill}
    \subfigure[``On a serene winter day, the country house stands gracefully amidst a pristine blanket of snow. "]{
      \includegraphics[width=0.21\textwidth]{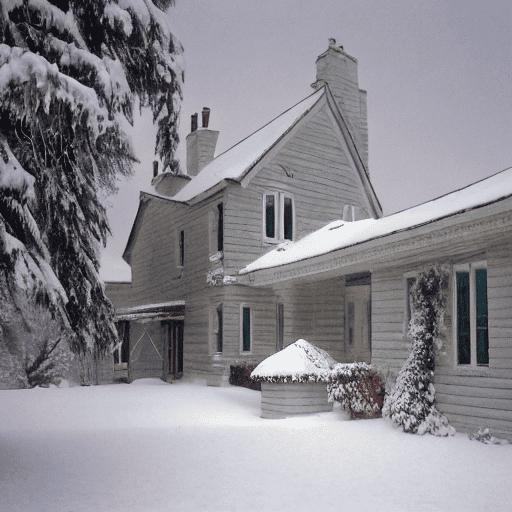}
    } \\
    \subfigure[``A wooden house built in the desert."]{
      \includegraphics[width=0.21\textwidth]{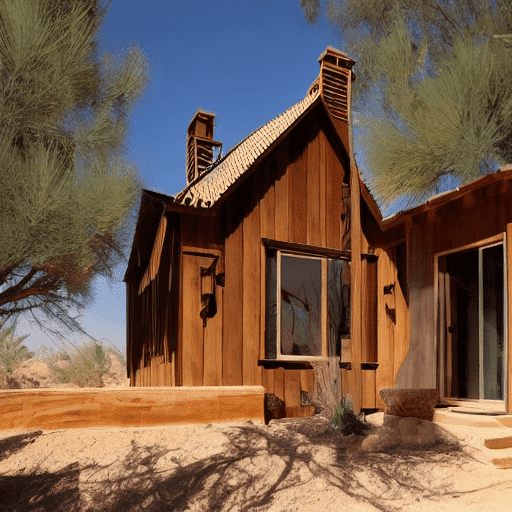}
    }
    \hspace*{\fill}
    \subfigure[``a house is built under the sea."]{
      \includegraphics[width=0.21\textwidth]{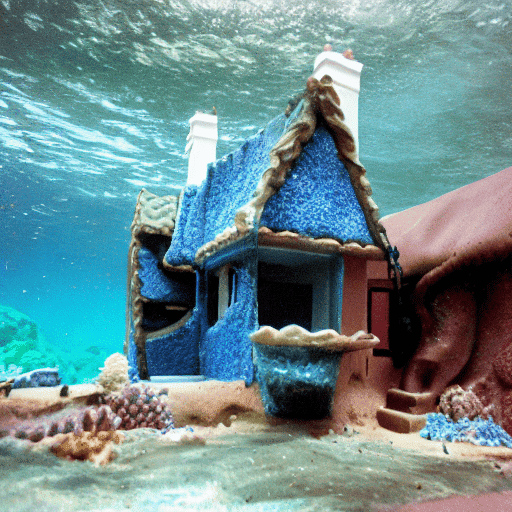}
    }
    \hspace*{\fill}
    \subfigure[``'' a country house is constructed with toy bricks."]{
      \includegraphics[width=0.21\textwidth]{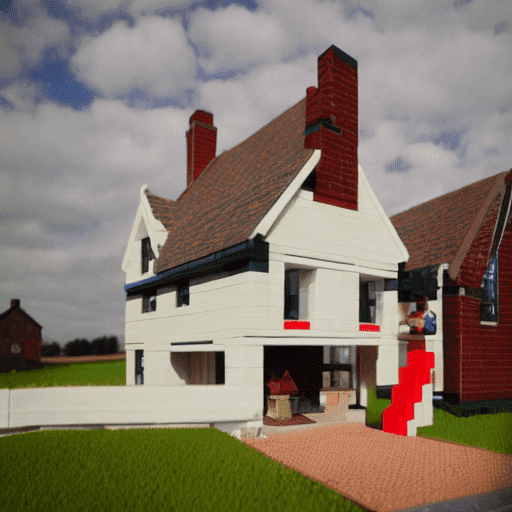}
    }
    \hspace*{\fill}
    \subfigure[``a House made of toy plasticine."]{
      \includegraphics[width=0.21\textwidth]{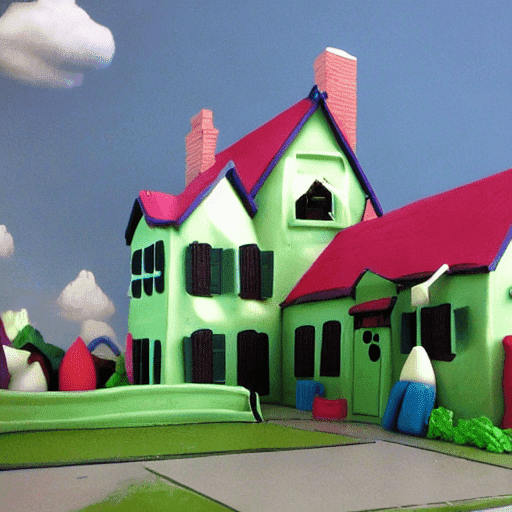}
    }
    \end{center}
    \label{prompt-fig-supp}
\end{figure}

The BPS model does not consider $C_{\text{text}}$ as an input. Investigating the integration of the BPS model with diverse textual prompts is necessary. The following example demonstrates the effective integration of inputs derived from various textual sources with the BPS model. Detailed textual descriptions are shown in Fig. \ref{prompt-fig-supp}.

Furthermore, regarding the generation tasks discussed in the paper, we provide the following generation samples. These experimental results demonstrate the BPS model's powerful generative capabilities and strong conditioning fidelity.

\begin{figure}[H]
    \vskip -0.1in
    \begin{center}
    \centerline{\includegraphics[width=0.9\textwidth]{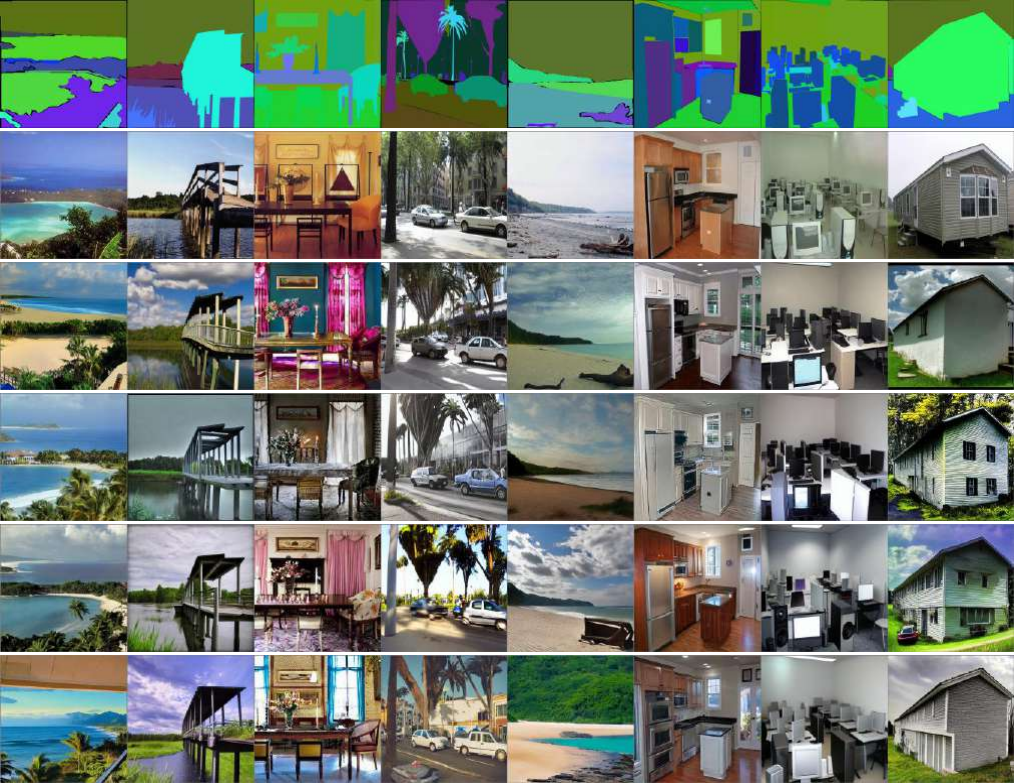}}
    \vskip -0.1in
    \caption{Segmentation-to-image. The first row is the condition, and the rest of the rows in each column are the corresponding generated structures}
    \label{seg-supp-fig}
    \end{center}
    \vskip -0.35in
\end{figure}

\begin{figure}[H]
    \vskip -0.1in
    \begin{center}
    \centerline{\includegraphics[width=0.9\textwidth]{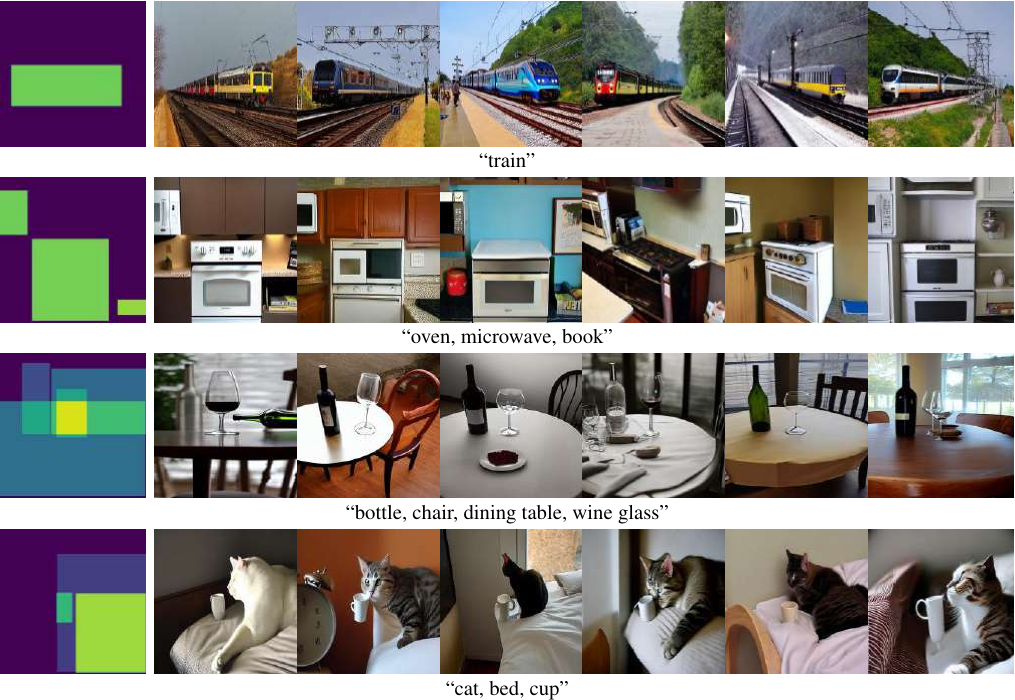}}
    \vskip -0.1in
    \caption{Layout-to-Image. The left column is the condition and the right column is the generated result.}
    \label{layout-supp-fig}
    \end{center}
    \vskip -0.35in
\end{figure}

\begin{figure}[H]
    \vskip -0.1in
    \begin{center}
    \centerline{\includegraphics[width=\textwidth]{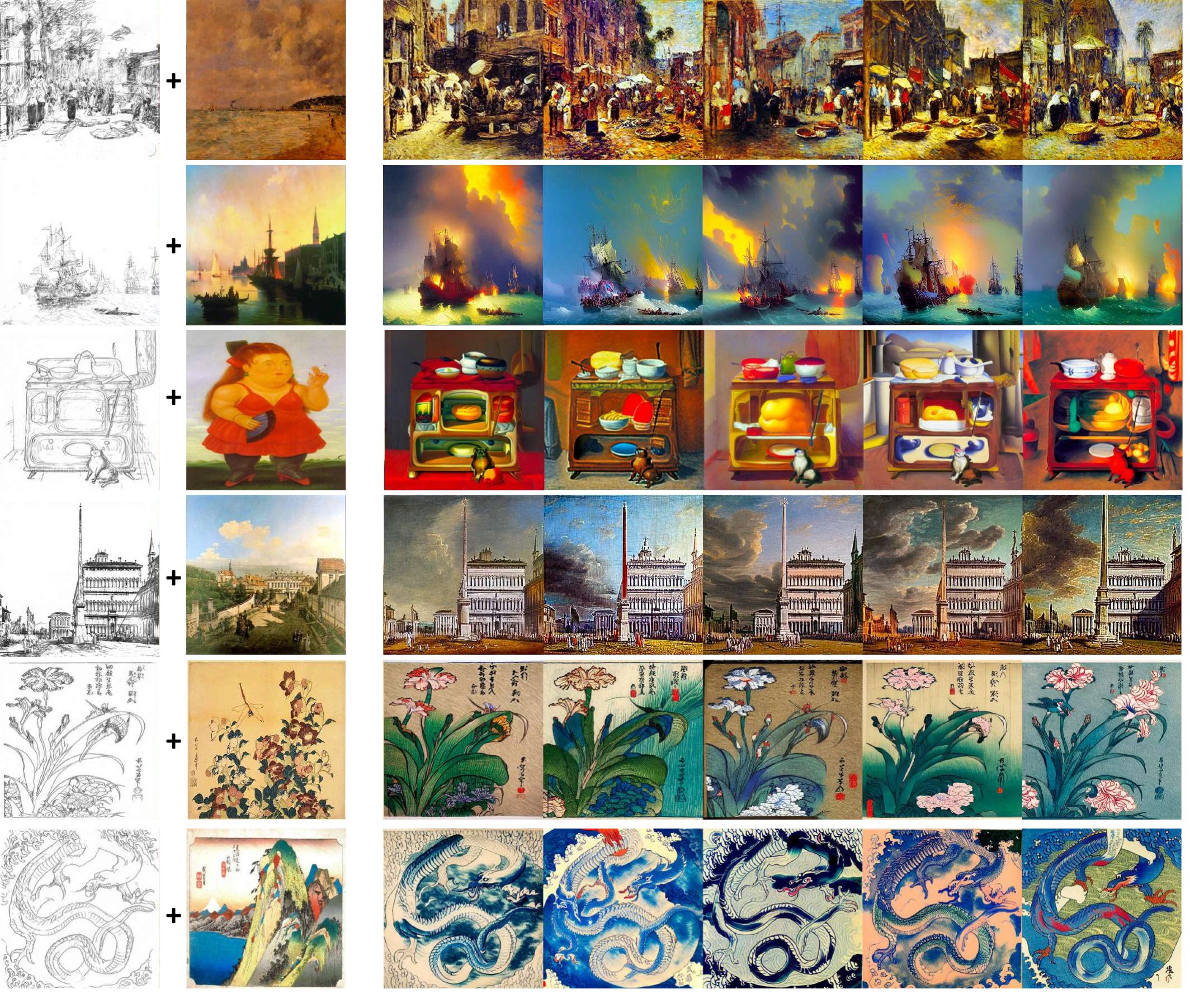}}
    \vskip -0.1in
    \caption{Artistic drawing. The left column is the line condition and the source image, the right column is the generated result.}
    \label{style-supp-fig}
    \end{center}
    \vskip -0.35in
\end{figure}

\begin{figure}[H]
    \vskip -0.1in
    \begin{center}
    \centerline{\includegraphics[width=\textwidth]{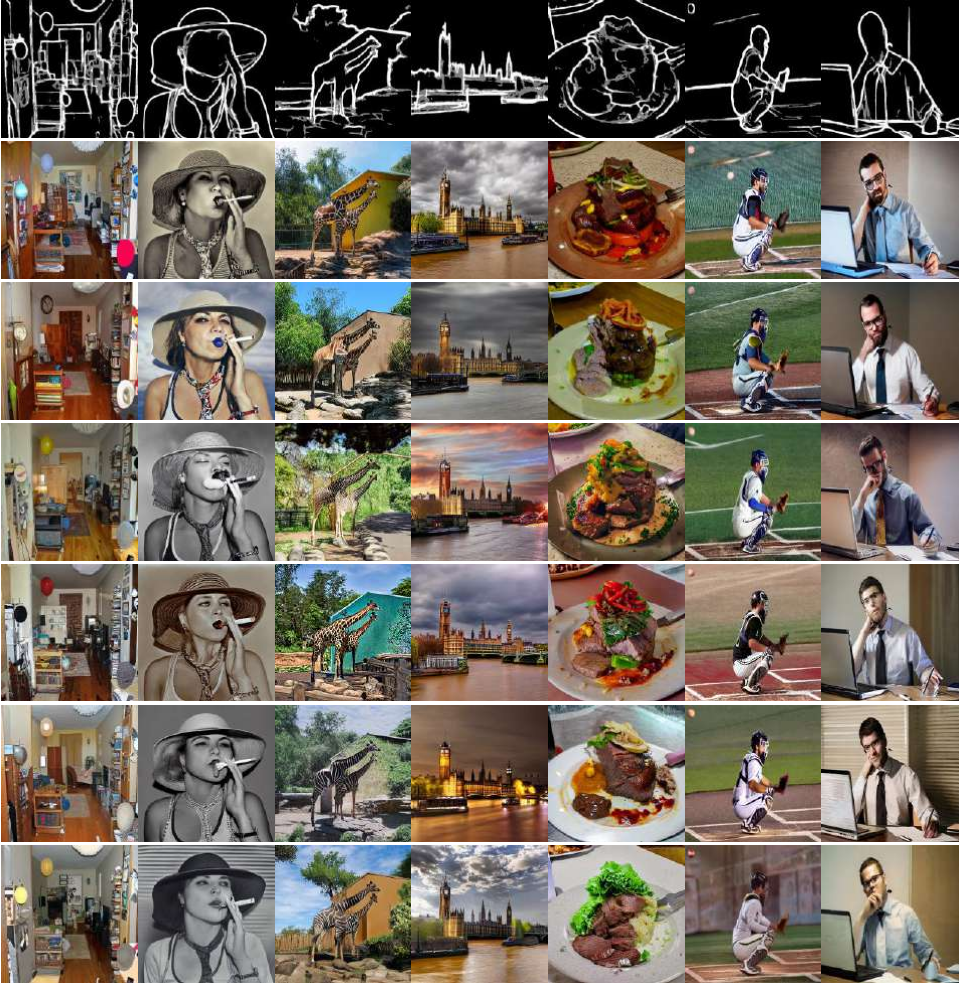}}
    \vskip -0.1in
    \caption{Sketch-to-image. The first row is the condition, and the rest of the rows in each column are the corresponding generated structures.}
    \label{sketch-supp-fig}
    \end{center}
    \vskip -0.35in
\end{figure}

\section{Comparison}
\begin{figure}[H]
    \vskip -0.1in
    \begin{center}
    \centerline{\includegraphics[width=\textwidth]{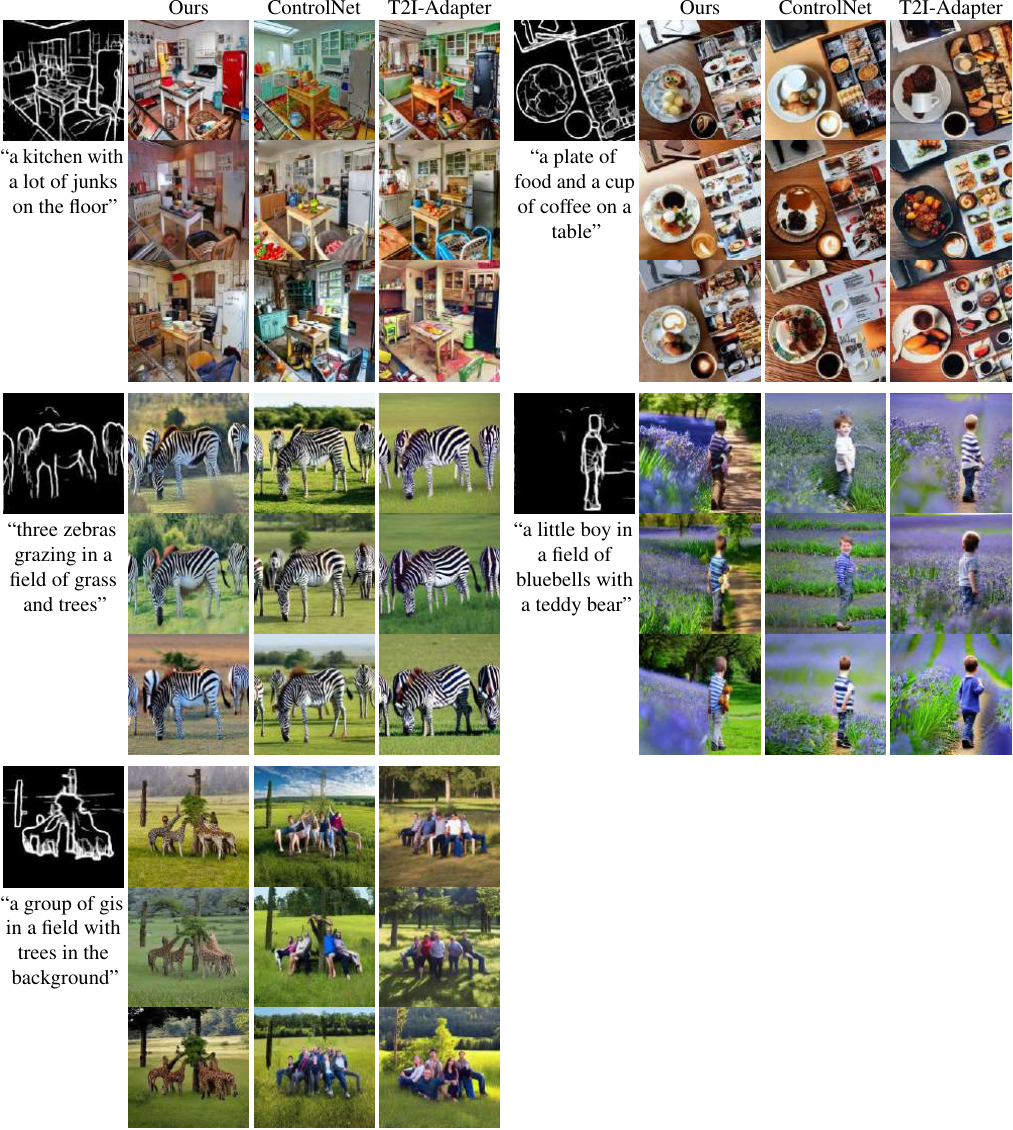}}
    \vskip -0.1in
    \caption{Comparison between our BPS, ControlNet \citep{zhang2023adding}, T2I-Adapter \citep{mou2023t2i}.}
    \label{compare-supp-fig}
    \end{center}
    \vskip -0.35in
\end{figure}

Fig. \ref{compare-supp-fig} presents a comparative analysis of the various methodologies employed in this study. The text data, generated using BLIP \citep{li2022blip}, aligns coherently with the main context of the research.


\section{Ablation Study}
\subsection{Data Efficiency}
This study utilized the coco17 dataset, with a random selection of data from the training and validation sets for model training. The models were then tested on the separate test set, and the generated results are presented below. Columns (a)-(e) in Fig. \ref{abla-supp-fig} provide a comparative analysis of the generation results obtained from models trained with datasets of different sizes.
\begin{figure}[H]
    \vskip -0.1in
    \begin{center}
    \centerline{\includegraphics[width=\textwidth]{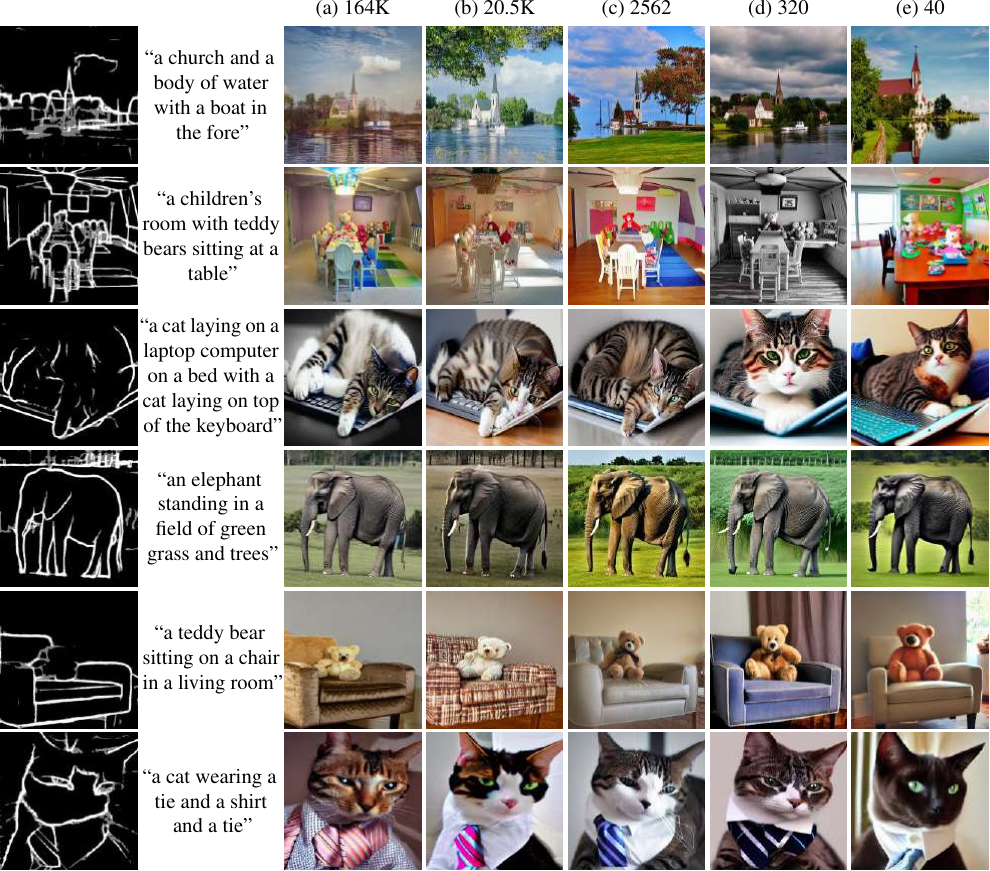}}
    \vskip -0.1in
    \caption{Comparative analysis of the generation results achieved by models trained using datasets of varying sizes.}
    \label{abla-supp-fig}
    \end{center}
    \vskip -0.35in
\end{figure}

\subsection{Intervention Strength}
This study examined the impact of varying intervention weights on control effects in two distinct tasks: the layout-to-image task and the sketch-to-image task.

\textbf{Layout-to-image}. In this experiment, the weights $w_i, i =1, ..., 21$ were uniformly initialized to 1. The effects generated under different weight settings, specifically 1, 0.8, and 0.5, are illustrated in columns (a), (b), and (c) of the figure below.

\begin{figure}[ht]
    \begin{center}
    \subfigure[Intervention weight settings and their impact on generation effects in the Layout-to-image task are depicted in the figure. Columns (a), (b), and (c) represent different intervention weights applied to the same BPS model.]{
      \includegraphics[width=0.4\textwidth]{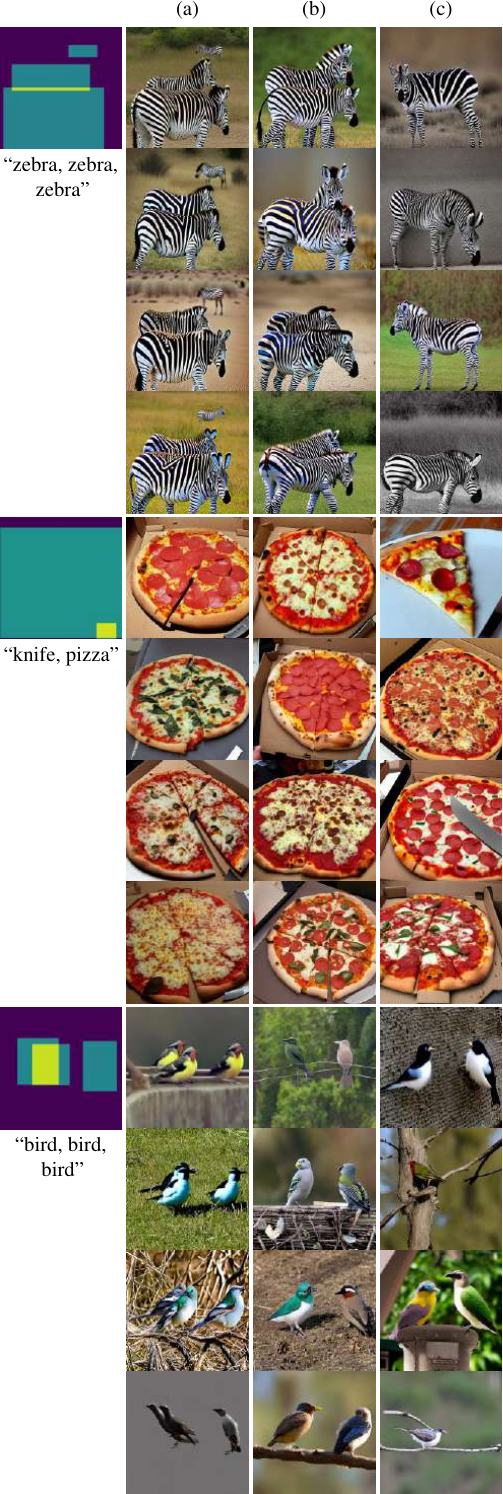}
    } \label{weight1-sup-fig}
    \subfigure[Intervention weight settings and their impact on generation effects in the sketch-to-image task are depicted in the figure. Columns (a), (b), and (c) represent different intervention weights applied to the same BPS model.]{
      \includegraphics[width=0.4\textwidth]{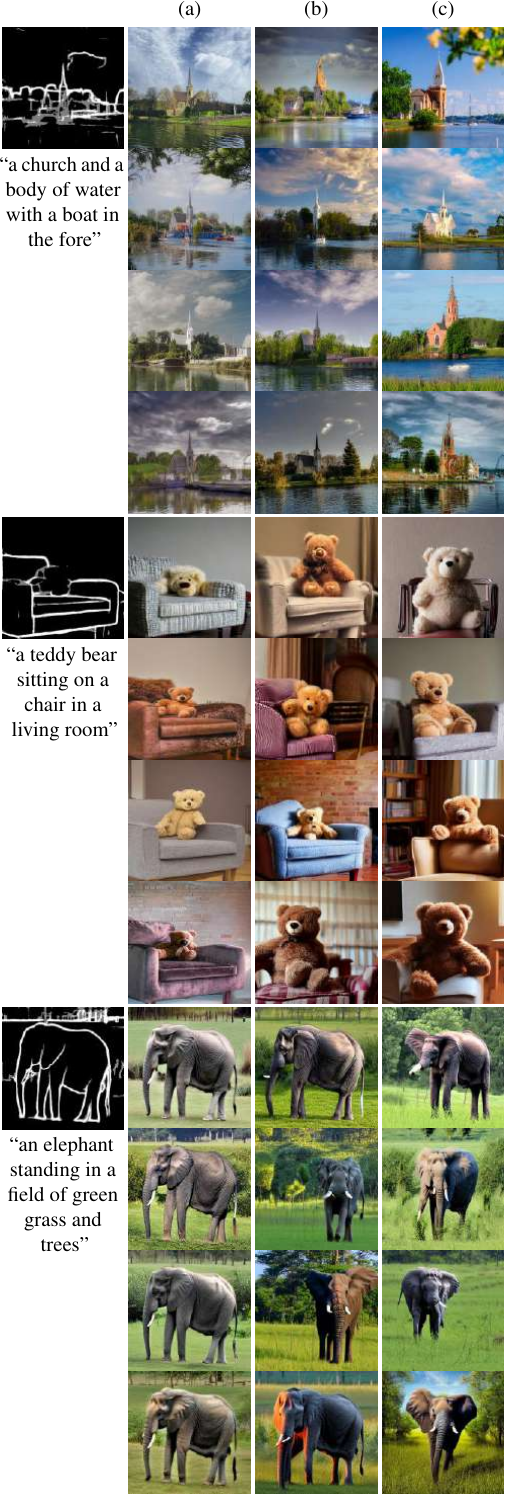}
    }
    \label{weight2-sup-fig}
    \end{center}
    \vspace{-0.1in}
\end{figure}

\textbf{Sketch-to-image}. In this experiment, the weights were initialized as $w_i = [16,16,8,8,4,4,2,2,16]$ for $i = 1, ..., 9$, and $w_i = 1$ for $i = 10, ..., 21$, in line with the configuration described in subsection \ref{sec-Differentiated-integration-structure}. Columns (a), (b), and (c) in Fig. \ref{weight2-sup-fig} demonstrate the down-weighting of intervention weights for the encoder and middle block to 1, 1/2, and 1/4, respectively.

\end{document}